\documentclass{article}

    \usepackage[final]{neurips_2024}

\usepackage[utf8]{inputenc} 
\usepackage[T1]{fontenc}    
\usepackage[colorlinks=true]{hyperref}  

\usepackage{url}            
\usepackage{booktabs}       
\usepackage{amsfonts}       
\usepackage{nicefrac}       
\usepackage{microtype}      
\usepackage{xcolor}         
\usepackage{enumitem}

\usepackage{amsmath}
\usepackage{amssymb}
\usepackage{mathtools}
\usepackage{amsthm}
\usepackage{xspace} 

\usepackage{colortbl}

\usepackage{subfigure}
\usepackage{graphicx}

\newcommand{\benchname}{I$^2$EBench\xspace}

\title{\benchname: A Comprehensive Benchmark for Instruction-based Image Editing}

\author{
    \textbf{Yiwei Ma}$^*$  \quad 
    \textbf{Jiayi Ji}\thanks{Equal contribution.} \quad 
    \textbf{Ke Ye} \quad 
    \textbf{Weihuang Lin} \quad 
    \textbf{Zhibin Wang} \\
    \textbf{Yonghan Zheng} \quad 
    \textbf{Qiang Zhou} \quad 
    \textbf{Xiaoshuai Sun}\thanks{Corresponding author.} \quad 
    \textbf{Rongrong Ji} \\
    \texttt{yiweima@stu.xmu.edu.cn} \quad
    \texttt{xssun@xmu.edu.cn}
}

\begin{document}

\maketitle

\begin{abstract}
    Significant progress has been made in the field of Instruction-based Image Editing (IIE). However, evaluating these models poses a significant challenge. 
    A crucial requirement in this field is the establishment of a comprehensive evaluation benchmark for accurately assessing editing results and providing valuable insights for its further development. 
    In response to this need, we propose \textbf{\benchname}, a comprehensive benchmark designed to automatically evaluate the quality of edited images produced by IIE models from multiple dimensions.
    \benchname consists of 2,000+ images for editing, along with 4,000+ corresponding original and diverse instructions. 
    It offers three distinctive characteristics:
    1) {\emph{Comprehensive Evaluation Dimensions:}}  \benchname comprises 16 evaluation dimensions that cover both high-level and low-level aspects, providing a comprehensive assessment of each IIE model.
    2) {\emph{Human Perception Alignment:}} To ensure the alignment of our benchmark with human perception, we conducted an extensive user study for each evaluation dimension.
    3) {\emph{Valuable Research Insights:}} By analyzing the advantages and disadvantages of existing IIE models across the 16 dimensions, we offer valuable research insights to guide future development in the field.
    We will open-source \benchname, including all instructions, input images, human annotations, edited images from all evaluated methods, and a simple script for evaluating the results from new IIE models.
    The code, dataset and generated images from all IIE models are provided in github: \textcolor{blue}{\url{https://github.com/cocoshe/I2EBench}}.

\end{abstract}

\section{Introduction}

\begin{figure}
  \centering
  \includegraphics[width=1.0\columnwidth]{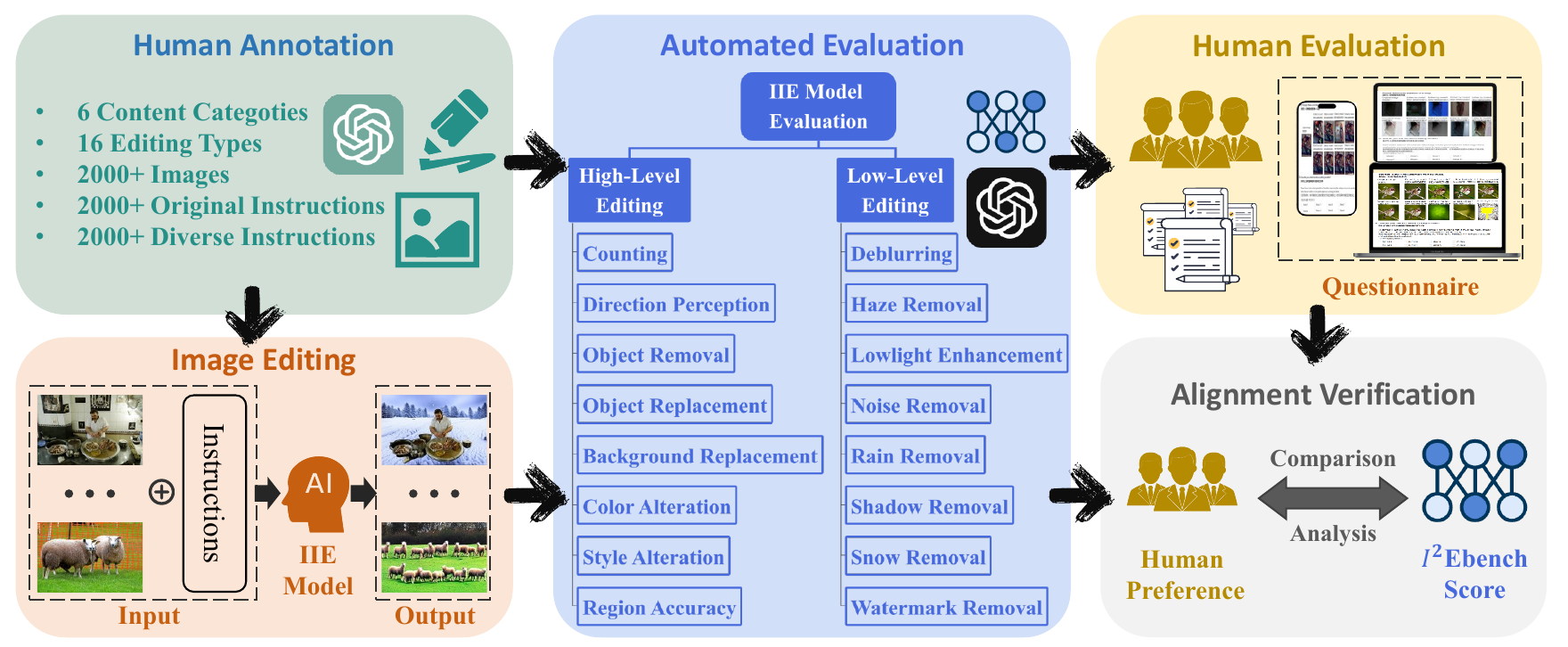}
  \caption{
  Overview of \benchname, an automated system for evaluating the quality of editing results generated by instruction-based image editing (IIE) models.
  We collected a dataset of over 2000+ images from public datasets~\cite{lin2014microsoft,guo2023sky,MartinFTM01,chen2021all,ancuti2019dense,liu2021synthetic,Liu_2021_WACV,qu2017deshadownet,Nah_2017_CVPR,shen2019human,wei2018deep} and annotated them with corresponding original editing instructions. To diversify the instructions, we used ChatGPT~\cite{achiam2023gpt} to generate varied versions.
  With the collected images and the original/diverse editing instructions, we utilized existing IIE models to generate edited images.
  Subsequently, we developed an evaluation methodology to automatically assess the adherence of edited images to the provided instructions under different dimensions.
  We also implemented human evaluation to obtain human preferences for editing results of different IIE models.
  Finally, we analyzed the correlation between automated evaluation and human evaluation, confirming alignment with human perception.
  }
  \label{fig:overview}
\end{figure}

Instruction-based Image Editing (IIE)\cite{brooks2023instructpix2pix,geng2023instructdiffusion,zhang2024magicbrush,li2023instructany2pix,wang2023instructedit,zhang2023hive,fu2023guiding}, which aims to edit an image using a text instruction, provides a user-friendly way for the community to edit images.
Over the past few years, significant progress has been made in IIE, with the development of diffusion models~\cite{ho2020denoising,sohl2015deep,welling2011bayesian,kulikov2023sinddm} and large vision-language models (LVLMs)~\cite{liu2023improvedllava,liu2023llava,fei2024enhancing,fei2024video,fei2024vitron,ma2024inf}. 
However, there is a pressing need for a comprehensive benchmark to effectively assess the performance of these models.
An ideal evaluation framework should not only measure the editing quality across different dimensions but also align with human perception to ensure reliable measurements.
Furthermore, the evaluation should highlight the specific strengths and weaknesses of each model, thereby offering valuable insights for future endeavors in data selection, training strategy selection, and architecture design within this field.
However, evaluating an IIE model poses challenges due to the diverse range of editing types and the inherent difficulty in assessing the level of alignment between edited images and given instructions.

Existing evaluation metrics for IIE could be divided into three categories: 1) conventional metric; 2) user study; 3) benchmark.
The first category~\cite{brooks2023instructpix2pix,geng2023instructdiffusion,zhang2024magicbrush,li2023instructany2pix,wang2023instructedit,huang2023smartedit} employs conventional metrics to evaluate IIE models, including CLIP Score~\cite{radford2021learning}, CLIP Text-Image Direction Similarity~\cite{radford2021learning}, PSNR~\cite{korhonen2012peak}, SSIM~\cite{wang2004image}, 
and LPIPS~\cite{zhang2018unreasonable}. The advantage of this approach is its ease of use. However, a single metric is not suitable for evaluating all types of editing. For instance, CLIP score measures the similarity between images and text, making it less suitable for low-level visual editing tasks like denoising and low-light enhancement. Similarly, PSNR, which measures image similarity, is not adequate for high-level visual editing tasks such as object removal and replacement.
The second category~\cite{li2023instructany2pix,zhang2023hive,fu2023guiding} involves methods that evaluate the effectiveness of different techniques by soliciting ratings from human participants. This approach directly reflects human preferences and aligns the results with human perception. However, it is a costly method and lacks reproducibility, as the test sets and participants may be not consistent in each evaluation. 
%
The final category comprises benchmarks~\cite{kawar2023imagic,wang2023imagen,basu2023editval,huang2024diffusion} specifically designed for evaluating IIE models. While these benchmarks are tailored for IIE, they have certain limitations. For example, TedBench~\cite{kawar2023imagic} evaluates only 100 images with commonly occurring editing types, which may not sufficiently demonstrate the capabilities of IIE models. EditBench~\cite{wang2023imagen} focuses on mask-guided editing, rendering it unsuitable for evaluating mask-free methods. In EditVal~\cite{basu2023editval}, only a limited set of dimensions related to size or location can be automatically evaluated, limiting its universality.

In this paper, we propose \benchname, a comprehensive benchmark designed to automatically evaluate the performance of IIE models.
\benchname exhibits three attractive characteristics: 1) Comprehensive Evaluation Dimension, 2) Human Perception Alignment, and 3) Valuable Research Insights.

First and foremost, \benchname offers a comprehensive evaluation dimension.
These dimensions are categorized into two main types: \textit{High-level Editing} and \textit{Low-level Editing}.
High-level editing primarily focuses on understanding instructions or editing specific areas of images, whereas low-level editing is more concerned with editing image details or the entire image.
As shown in Fig.~\ref{fig:overview}, both high-level and low-level editing consist of 8 fine-grained editing dimensions, which serve to demonstrate the model's proficiency in high-level and low-level editing.
%
%
%
We meticulously collected approximately 140 images for each editing dimension and annotated each image with an original editing text instruction. 
To diversify the instructions, we also utilized ChatGPT~\cite{achiam2023gpt} to enhance the description of the instructions and obtain a wider range of variations.
In addition to the multi-dimensional evaluation, we also conducted a multi-category evaluation to assess the model's performance on different content categories. 
To achieve this, we included additional annotations for each instruction with different categories such as Animal, Object, Scenery, Plant, Human, and Global.

\begin{figure}
  \centering
  \includegraphics[width=1.0\columnwidth]{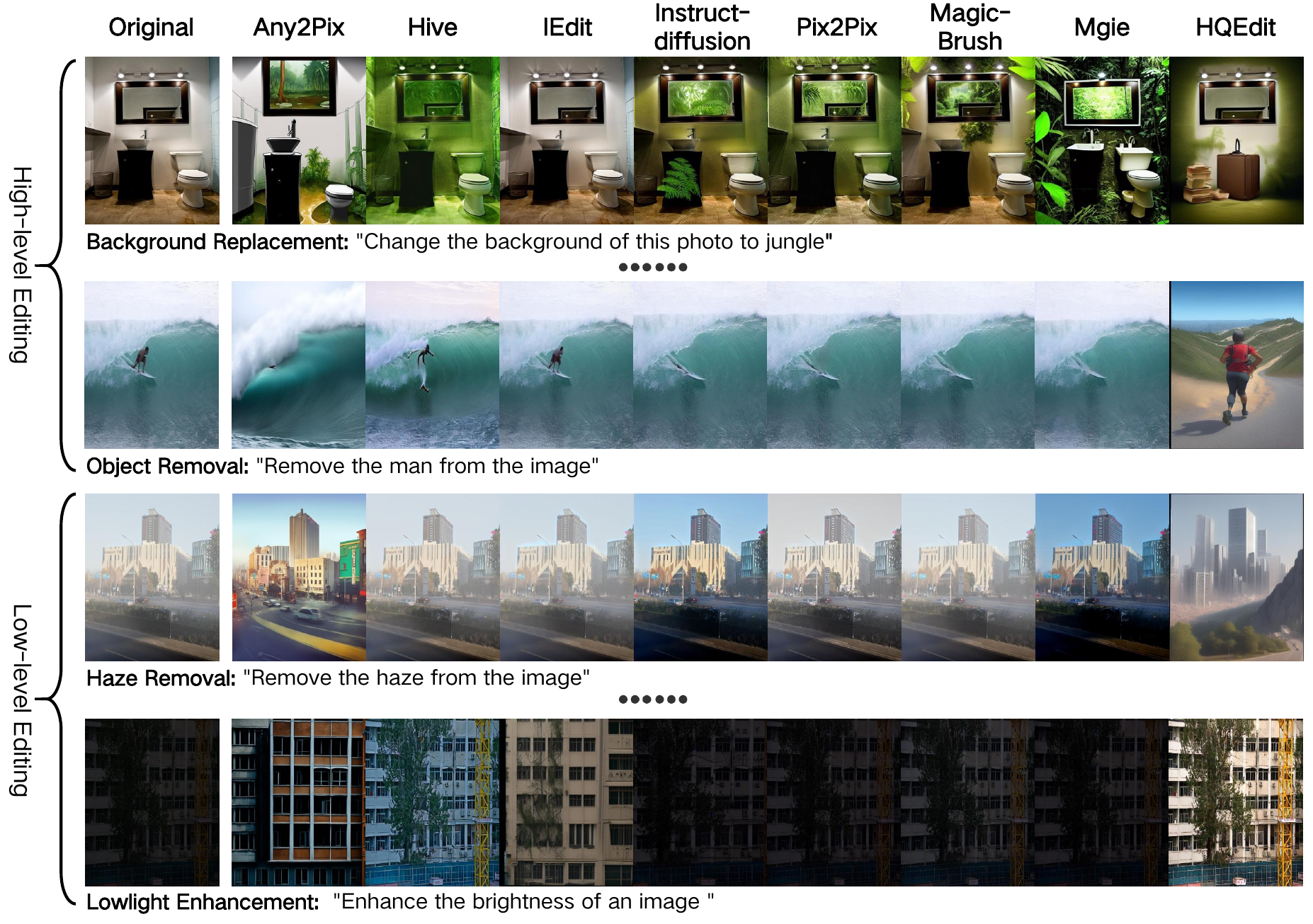}
  \caption{
  Visualization of the editing results on the proposed 16 evaluation dimensions using different IIE models, including InstructAny2Pix~\cite{li2023instructany2pix}, HIVE~\cite{zhang2023hive}, InstructEdit~\cite{wang2023instructedit}, InstructDiffusion~\cite{geng2023instructdiffusion}, InstructPix2Pix~\cite{brooks2023instructpix2pix}, MagicBrush~\cite{zhang2024magicbrush}, MGIE~\cite{fu2023guiding}, and HQEdit~\cite{hui2024hq}. A detailed version can be found in supplementary materials.
  }
  \label{fig:res_sample}
\end{figure}

Second, the \benchname score aligns with human perception.
This is accomplished by collecting scores from human annotators for the outputs generated by different IIE models, covering multiple evaluation dimensions. By conducting a comprehensive analysis of both the \benchname scores and the human scores, we have identified a substantial correlation between them. This discovery serves as compelling evidence, affirming that our proposed evaluation approach closely aligns with human perception.

Lastly, \benchname offers valuable research insights through its systematic evaluation across various dimensions and categories. The proposed \benchname not only facilitates a comprehensive assessment of existing models but also derives valuable insights into their respective strengths and weaknesses. These insights act as a roadmap for enhancing architecture design, refining data selection strategies, and ultimately elevating the quality of editing outcomes.

We are open-sourcing \benchname, including all instructions, input images, human annotations, edited images from all evaluated methods (like Fig.~\ref{fig:res_sample}), and a simple script for evaluating the results of new IIE models. By making these resources freely available, we aim to foster fair comparisons within the field and facilitate valuable insights for community development.

\section{Related Work}


\subsection{Instruction-based Image Editing}
With the advancements in Generative Adversarial Networks (GAN)\cite{goodfellow2014generative,goodfellow2020generative,mao2017least,karras2019style,yoon2019time,karras2020training,chen2018cartoongan,zhang2019self} and Diffusion models\cite{song2020denoising,ho2020denoising,nichol2021improved,kawar2022denoising,austin2021structured,dockhorn2022genie}, text-to-image models~\cite{saharia2022photorealistic,rombach2022high,ramesh2021zero,ramesh2022hierarchical,betker2023improving,karras2019style,karras2020analyzing} have made remarkable progress in recent years.
As the demand for image editing continues to grow, a multitude of text-based image editing~\cite{xu2024cyclenet,kawar2023imagic,zhang2023sine,saund2003perceptually,zhang2024text} models have emerged.
One editing task, known as Prompt-based Image Editing (PIE)~\cite{avrahami2022blended,valevski2023unitune,hertz2022prompt,dong2023prompt}, requires users to provide a target description along with the original image. The PIE model then analyzes the target description to modify the input image accordingly, generating a target image that matches the provided description.
However, despite the lowered threshold for image editing, the requirement of describing the entire content of the target image in the description still poses challenges in terms of user interaction.
To address this limitation, Instruction-based Image Editing (IIE)~\cite{brooks2023instructpix2pix,geng2023instructdiffusion,li2023instructany2pix,fu2023guiding,huang2023smartedit} was proposed, which simplifies the user's role to providing the original image and modification instructions (\emph{e.g.,} `Remove the dog').
One notable implementation, InstructPix2Pix~\cite{brooks2023instructpix2pix}, introduces a large-scale dataset for instruction-based image editing. The dataset is created using a fine-tuned GPT-3~\cite{brown2020language} and image pairs generated by the Prompt-to-Prompt diffusion model~\cite{hertz2022prompt}. Additionally, InstructPix2Pix proposes an instruction-based diffusion model for image editing based on this dataset.
However, due to the automatic generation and filtering of the InstructPix2Pix dataset, concerns arise regarding its quality and potential noise. To address this, MagicBrush~\cite{zhang2024magicbrush} proposes a manually-annotated instruction-guided image editing dataset.
In addition to textual instructions, InstructAny2Pix~\cite{li2023instructany2pix} proposes a model that utilizes other modalities, such as audio and image, as instructions.
To enhance the level of detail in instructions and improve the accuracy of editing results, MGIE~\cite{fu2023guiding} introduces the use of Multimodal Large Language Models (MLLM)~\cite{liu2023improvedllava}.
SmartEdit~\cite{achiam2023gpt}, aiming to improve the editing capabilities of IIE models in complex scenes, incorporates MLLM into the IIE model to better comprehend instructions. 
%
Despite significant progress, evaluating the editing performance of IIE models remains a crucial concern. Therefore, in this paper, we present \benchname, a systematic evaluation framework for these models. Our work includes an in-depth analysis of their strengths and weaknesses, offering valuable insights for the future development of IIE models.

\subsection{Text-based Image Editing Benchmark}
While numerous benchmarks~\cite{marino2019ok,hudson2019gqa,bigham2010vizwiz,lu2022learn,li2023evaluating,li2023seed,li2023mvbench,yu2023mm,wu2023q} have been introduced for evaluating vision-language tasks~\cite{alayrac2022flamingo,li2023blip,ye2023mplug,wu2023visual,dai2024instructblip,hu2024mplug}, the evaluation of text-based image editing models often relies on metrics such as CLIP Score~\cite{radford2021learning}, PSNR~\cite{korhonen2012peak}, SSIM~\cite{wang2004image}, and LPIPS~\cite{zhang2018unreasonable}.
Several existing studies have introduced benchmarks to assess the performance of image editing models.
TedBench~\cite{kawar2023imagic} presents a relatively small benchmark consisting of only 100 images and a limited set of highly common editing types.
EditBench~\cite{wang2023imagen} is specifically designed to evaluate mask-guided image editing methods, which necessitate the availability of additional masks indicating the areas to be edited.
In EditVal~\cite{basu2023editval}, the evaluation of certain dimensions relies on manual labor, thereby limiting the reproducibility of performance. Moreover, the remaining dimensions primarily involve modifications to object size or position, lacking comprehensive coverage.
While MagicBrush~\cite{zhang2024magicbrush} and Emu Edit~\cite{sheynin2023emu} propose test sets for evaluating editing performance, they still rely on conventional metrics such as L1, L2, CLIP-I, DINO, and CLIP-T, which may not accurately capture the nuances of all editing types.
SmartEdit~\cite{huang2023smartedit} specifically develops a benchmark tailored for complex editing scenarios, but it does not accommodate other editing scenarios.
Considering the current absence of a systematic benchmark that comprehensively evaluates the editing performance of IIE models across different editing types, we propose \benchname to address this gap.

\section{\texorpdfstring{I$^2$EBench}{I2EBench}}

This section provides an overview of the main components of \benchname.
In Sec.~\ref{sec:eval}, we provide a concise introduction to the principles, definitions, and evaluation methods of 16 dimensions. 
Sec.~\ref{sec:annotation} outlines the process of data annotation.
Lastly, in Sec.~\ref{sec:human}, we present the human evaluation process to assess the correlation between the \benchname score and the human score.
\textit{A detailed explanation can be found in the supplementary materials.}

\subsection{Evaluation Dimension}
\label{sec:eval}

In our evaluation of the IIE model's editing quality, we have categorized it into 16 dimensions, each assessing different aspects of editing in a top-down manner.
An overview of \benchname is presented in Fig.~\ref{fig:overview}.
{High-level Editing Evaluation} primarily focuses on assessing the model's ability to accurately understand instructions and make precise edits to local areas of the input image. This evaluation consists of 8 dimensions.
{Low-level Editing Evaluation}, on the other hand, primarily examines global editing and detailed image processing. It also comprises 8 evaluation dimensions. 
Unlike previous approaches~\cite{fu2023guiding, zhang2023hive, geng2023instructdiffusion} that relied on a single metric, such as CLIP score~\cite{radford2021learning}, to evaluate editing quality for all editing types, we have developed specialized evaluation methods for each of the 16 dimensions.
This approach is necessary due to the distinct goals of high-level and low-level editing. 

\subsubsection{High-level Editing}
Evaluating editing quality in high-level dimensions poses a challenge due to the diverse goals involved, making it impractical to rely on a single metric.
The advancement of Multimodal Large Language Models (MLLM)~\cite{gao2024sphinx,chu2024mobilevlm,zhu2024vislinginstruct,dong2024internlm,ma2022x,ma2023towards,ji2022knowing}, such as GPT-4V~\cite{achiam2023gpt}, Gemini Pro~\cite{reid2024gemini}, and QWen-VL-Plus~\cite{bai2023qwen}, has significantly enhanced automated understanding of images.
Therefore, to ensure precise evaluation of the editing quality of IIE models in high-level dimensions, we leverage the exceptional capabilities of the widely recognized GPT-4V model to make judgments for most high-level evaluation dimensions.

\noindent{\textbf{Counting.}} 
The Counting dimension pertains to instructions related to the number of objects, such as "add two apples to the image."
To assess this dimension, we query GPT-4V about the number of target objects in the image and compare its response with the human-annotated answer.

\noindent{\textbf{Direction Perception.}} 
The Direction Perception dimension requires the IIE model to comprehend directions provided in instructions, and accurately make edits when presented with images.
We evaluate this dimension by asking GPT-4V if the target object is in the expected position.

\noindent{\textbf{Object Removal.}} 
The Object Removal dimension focuses on removing the target object according to the given instruction.
To evaluate this dimension, we inquire whether GPT-4V identifies the presence of the target object in the image.

\noindent{\textbf{Object Replacement.}} 
The Object Replacement dimension aims to replace the original object with the target object as instructed.
To assess this dimension, we query GPT-4V about the presence of the target object in the image.

\noindent{\textbf{Background Replacement.}} 
The Background Replacement dimension involves replacing the original background with the target background as specified in the instruction.
To evaluate this dimension, we ask GPT-4V if the background of the image matches the textual instruction.

\noindent{\textbf{Color Alteration.}} 
In the Color Alteration dimension, we modify the color of the target object using instructions.
To evaluate this dimension, we inquire  GPT-4V  about the color of the target object in the edited image.

\noindent{\textbf{Style Alteration.}} 
The Style Alteration dimension focuses on changing the style of the image.
To evaluate this dimension, we calculate the CLIP similarity~\cite{radford2021learning} between the edited image and "an image with $\cdots$ style".

\noindent{\textbf{Region Accuracy.}}  
In the editing task, we not only assess whether the target area has been edited correctly but also whether areas that should not be edited have been altered.
To evaluate this dimension, we sample input images and instructions from the Object Removal, Object Replacement, and Color Alteration dimensions.
We annotate the mask for the area that requires editing.
Next, we fill the mask area of the images before and after editing with white and calculate SSIM~\cite{wang2004image} to evaluate this dimension.

\begin{figure}
  \centering
  \includegraphics[width=1.0\columnwidth]{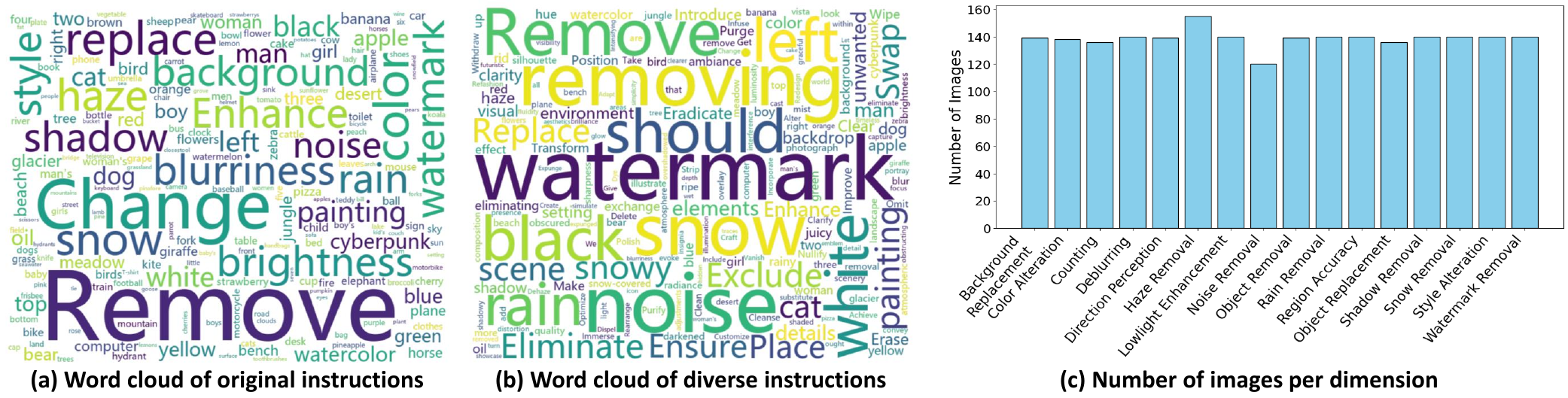}
  \caption{
    Word cloud visualization (a,b) and image quantity statistics (c) of \benchname.
  }
  \label{fig:data}
\end{figure}

\subsubsection{Low-level Editing}
Unlike high-level editing, low-level editing instructions are simpler, lacking specifications regarding object size, orientation, or color.
Various low-level editing tasks~\cite{wang2023omni,chen2023masked,sanghvi2023structured,chen2023learning,wu2023ridcp,guo2023shadowdiffusion,kong2022reflash} have undergone extensive development over the years, resulting in a relatively mature evaluation system. Therefore, for low-level editing, we employ the widely recognized metric, namely SSIM~\cite{wang2004image}, to evaluate the editing quality.

\noindent{\textbf{Deblurring.}} 
Deblurring encompasses the procedure of mitigating or eliminating blur from images, resulting in enhanced clarity and sharpness. 

\noindent{\textbf{Haze Removal.}}
Haze removal entails the elimination or reduction of atmospheric haze or fog from images, augmenting visibility and reinstating the true colors and intricate details of the scene. 

\noindent{\textbf{Lowlight Enhancement.}} 
Lowlight enhancement refers to the process of improving the quality of images captured in low-light conditions, enhancing brightness, and reducing noise.

\noindent{\textbf{Noise Removal.}} 
Noise removal involves the reduction or elimination of unwanted noises in images, resulting in cleaner and more visually appealing visuals.

\noindent{\textbf{Rain Removal.}} 
Rain removal aims to eliminate or reduce the visual effects of raindrops or rain streaks from images, improving clarity and restoring the original appearance.

\noindent{\textbf{Shadow Removal.}} 
Shadow removal refers to reducing or eliminating unwanted shadows from images, enhancing visibility, and improving overall image quality.

\noindent{\textbf{Snow Removal.}} 
The goal of Snow Removal is to effectively reduce or eliminate snow from images.

\noindent{\textbf{Watermark Removal.}} 
Watermark removal involves the removal or elimination of embedded watermarks from images, restoring the original appearance without the presence of the watermark.

\subsection{Human Annotation}
\label{sec:annotation}

\noindent{\textbf{Data Annotation}.}
We meticulously curated approximately 140 images from publicly available datasets~\cite{lin2014microsoft, guo2023sky, MartinFTM01, chen2021all, ancuti2019dense, liu2021synthetic, Liu_2021_WACV, qu2017deshadownet, Nah_2017_CVPR, shen2019human, wei2018deep} for each evaluation dimension of \benchname. The distribution of the image count for each dimension is illustrated in Fig.~\ref{fig:data}~(c). These images were then meticulously annotated with textual editing instructions by human annotators, namely original instructions. However, instructions provided by human annotators usually followed a singular sentence pattern. For instance, the prevalent sentence pattern for the object removal dimension was typically "remove $\cdots$ from the image". To foster increased diversity, we employed ChatGPT~\cite{achiam2023gpt} to effectively rewrite the original instructions.
Fig.~\ref{fig:data} (a) and (b) present the word cloud visualizations of the original and diverse instructions, respectively.
Additionally, we also annotate a category for each instruction, such as animal, object, scenery, plant, human, and global.

\begin{figure}
  \centering
  \includegraphics[width=1.0\columnwidth]{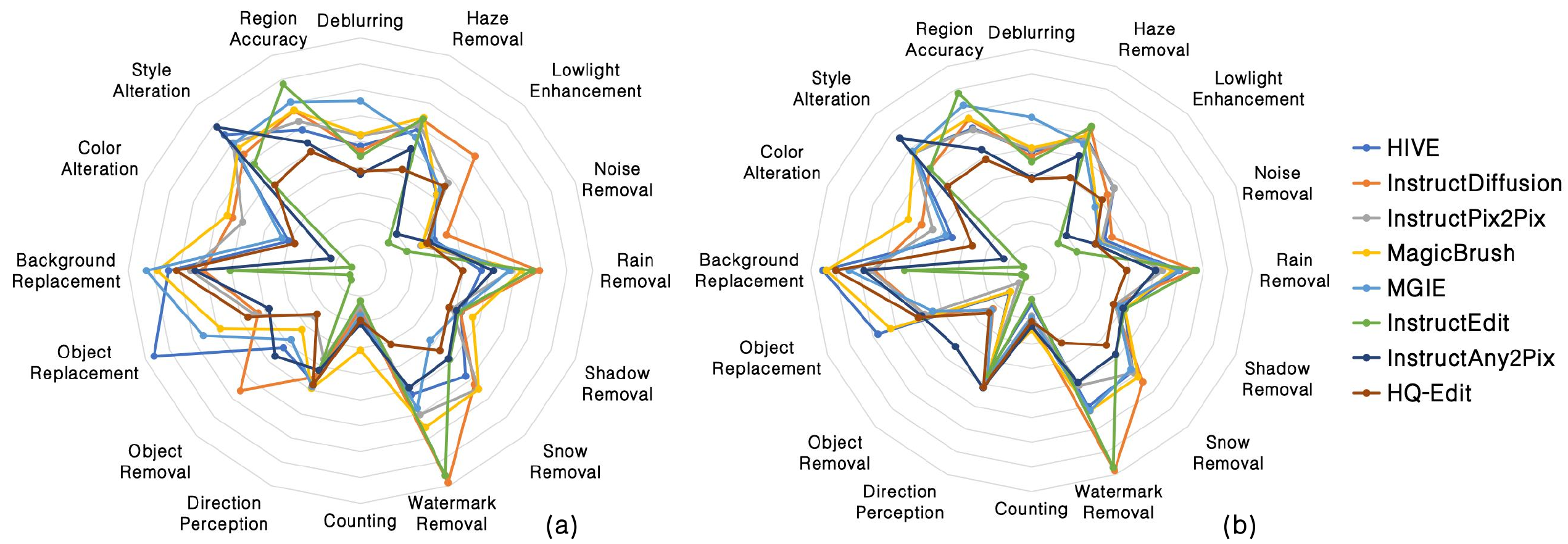}
  \caption{
    Comparison of radar charts for \benchname scores in different dimensions using (a) original instructions and (b) diverse instructions.
  }
  \label{fig:score_radar}
\end{figure}

\noindent{\textbf{Evaluation Annotation}.}
The evaluation process for \benchname encompasses two distinct categories.
The first category employs conventional metrics to assess various dimensions. For style alteration dimension, we utilize the CLIP score as a standard metric, which doesn't need any additional evaluation annotations. 
%
In the second category, we utilize GPT-4V to evaluate the quality of editing. To facilitate this evaluation, we enlisted the expertise of human annotators to annotate questions specifically designed for GPT-4V, along with corresponding standard answers.
For instance, let's consider the counting dimension and the instruction "Add a cat to the shoe rack". In this particular case, the annotated question provided by the human annotators is "How many cats are there on the shoe rack?", and the corresponding annotated answer is "One".

\subsection{Human Evaluation}
\label{sec:human}
The primary objective of the human evaluation is to ascertain the correlation between human perception and the \benchname score.
To achieve this, we present human evaluators with a textual instruction $T$, an input image $V_I$, and a set of edited images $\{V_1, V_2, \cdots, V_M\}$ generated by $M$ different IIE models. The evaluators are then tasked with ranking the results based on their judgment.
More specifically, we sample $N$ images for each evaluation dimension,  leading to a comprehensive collection of $N \times 16 \times 2$ edited image comparisons. Within each comparison, evaluators are presented with $M$ edited images to assess and rank in relation to one another.
We assign a human score to each model based on its ranking among the $M$ models.
Specifically, the model ranked first among the $M$ models receives a human score of $M$, while the model ranked last among the $M$ models receives a human score of 1. Additionally, the model ranked $k$ among the $M$ models is assigned a human score of $M - k + 1$.
To determine the human score for each dimension, we calculate the average of the human scores across all samples within that dimension.
Thus, the human score for each model ranges from 1 to $M$.

\begin{table}[]
\caption{\benchname evaluation results per dimension using original instructions. \textit{Exp Min} and \textit{Exp Max} denote the minimum and maximum values of all samples for each evaluation dimension.}
\setlength{\tabcolsep}{2pt}
\resizebox{\linewidth}{!}{
\begin{tabular}{l|c|c|c|c|c|c|c|c}
\midrule
\multicolumn{9}{c}{\textbf{Low-level Editing}} \\ 
\midrule
\textbf{Model}             & \textbf{Deblurring} & \textbf{\begin{tabular}[c]{@{}c@{}}Haze\\ Removal\end{tabular}}         & \textbf{\begin{tabular}[c]{@{}c@{}}Lowlight\\ Enhancement\end{tabular}} & \textbf{\begin{tabular}[c]{@{}c@{}}Noise\\ Removal\end{tabular}}      & \textbf{\begin{tabular}[c]{@{}c@{}}Rain\\ Removal\end{tabular}}           & \textbf{\begin{tabular}[c]{@{}c@{}}Shadow\\ Removal\end{tabular}}   & \textbf{\begin{tabular}[c]{@{}c@{}}Snow\\ Removal\end{tabular}}     & \textbf{\begin{tabular}[c]{@{}c@{}}Watermark\\ Removal\end{tabular}} \\
\midrule
\textbf{HIVE~\cite{zhang2023hive}}              & 44.25               & 54.89                                                                   & 37.61                                                                   & 24.59                                                                 & 45.47                                                                     & 37.61                                                               & 51.49                                                               & 49.99                                                                \\
\textbf{InstructDiffusion~\cite{geng2023instructdiffusion}} & 42.48               & 58.45                                                                   & \textbf{56.61}                                                          & \textbf{28.60}                                                        & \textbf{67.20}                                                            & 37.43                                                               & 55.65                                                               & \textbf{85.49}                                                       \\
\textbf{InstructPix2Pix~\cite{brooks2023instructpix2pix}}   & 48.03               & 56.15                                                                   & 43.32                                                                   & 20.11                                                                 & 56.64                                                                     & 34.19                                                               & 57.59                                                               & 58.12                                                                \\
\textbf{MagicBrush~\cite{zhang2024magicbrush}}        & 48.38               & \textbf{59.46}                                                          & 37.71                                                                   & 20.59                                                                 & 60.60                                                                     & \textbf{41.91}                                                               & \textbf{57.81}                                                      & 63.33                                                                \\
\textbf{MGIE~\cite{fu2023guiding}}              & \textbf{60.30}      & 51.75                                                                   & 39.99                                                                   & 23.25                                                                 & 56.00                                                                     & 36.91                                                               & 34.04                                                               & 55.53                                                                \\
\textbf{InstructEdit~\cite{wang2023instructedit}}      & 40.77               & 58.85                                                                   & 13.83                                                                   & 15.40                                                                 & 64.44                                                                     & 36.88                                                      & 43.45                                                               & 82.68                                                                \\
\textbf{InstructAny2Pix~\cite{li2023instructany2pix}}   & 34.34               & 47.27                                                                   & 18.03                                                                   & 22.89                                                                 & 49.94                                                                     & 35.84                                                               & 42.97                                                               & 47.28                                                                \\
\textbf{HQ-Edit~\cite{hui2024hq}} & 35.27 & 	39.25 & 	41.71 & 	22.13 & 	38.52 & 	33.13 & 	38.97 & 	29.80 
\\
\hline
\rowcolor[HTML]{F2F2F2} 
\textbf{Exp Min}           & 13.79               & 12.66                                                                   & 0.09                                                                    & 0.79                                                                  & 7.38                                                                      & 1.05                                                                & 2.18                                                                & 1.34                                                                 \\
\rowcolor[HTML]{F2F2F2} 
\textbf{Exp Max}           & 91.94               & 92.70                                                                   & 89.60                                                                   & 77.00                                                                 & 96.11                                                                     & 89.19                                                               & 89.26                                                               & 96.42                                                                \\
\midrule
\multicolumn{9}{c}{\textbf{High-level Editing}} \\
\midrule
\textbf{Model}             & \textbf{Counting}   & \textbf{\begin{tabular}[c]{@{}c@{}}Direction\\ Perception\end{tabular}} & \textbf{\begin{tabular}[c]{@{}c@{}}Object\\ Removal\end{tabular}}       & \textbf{\begin{tabular}[c]{@{}c@{}}Object\\ Replacement\end{tabular}} & \textbf{\begin{tabular}[c]{@{}c@{}}Background\\ Replacement\end{tabular}} & \textbf{\begin{tabular}[c]{@{}c@{}}Color\\ Alteration\end{tabular}} & \textbf{\begin{tabular}[c]{@{}c@{}}Style\\ Alteration\end{tabular}} & \textbf{\begin{tabular}[c]{@{}c@{}}Region\\ Accuracy\end{tabular}}   \\
\midrule
\textbf{HIVE~\cite{zhang2023hive}}              & 18.57               & 47.14                                                                   & 42.14                                                                   & \textbf{86.43}                                                        & 74.29                                                                     & 30.00                                                               & 25.32                                                               & 58.15                                                                \\
\textbf{InstructDiffusion~\cite{geng2023instructdiffusion}} & 15.00               & 44.29                                                                   & \textbf{65.71}                                                          & 42.86                                                                 & 60.71                                                                     & 53.57                                                               & 21.69                                                               & 66.18                                                                \\
\textbf{InstructPix2Pix~\cite{brooks2023instructpix2pix}}   & 13.57               & 37.14                                                                   & 25.00                                                                   & 44.29                                                                 & 65.71                                                                     & 49.29                                                               & 23.76                                                               & 61.63                                                                \\
\textbf{MagicBrush~\cite{zhang2024magicbrush}}        & \textbf{30.71}      & \textbf{49.29}                                                          & 32.14                                                                   & 58.57                                                                 & 78.57                                                            & \textbf{55.71}                                                      & 22.78                                                               & 66.34                                                                \\
\textbf{MGIE~\cite{fu2023guiding}}              & 17.14               & 48.57                                                                   & 37.86                                                                   & 65.71                                                                 & \textbf{82.86}                                                                     & 32.86                                                               & 23.68                                                               & 69.60                                                                \\
\textbf{InstructEdit~\cite{wang2023instructedit}}      & 11.76               & 41.73                                                                   & 5.04                                                                    & 4.41                                                                  & 50.36                                                                     & 3.62                                                                & 19.83                                                               & \textbf{77.08}                                                       \\
\textbf{InstructAny2Pix~\cite{li2023instructany2pix}}   & 20.59               & 41.73                                                                   & 46.76                                                                   & 38.24                                                                 & 64.03                                                                     & 12.32                                                               & \textbf{26.76}                                                      & 52.75                                                                \\
\textbf{HQ-Edit~\cite{hui2024hq}} & 19.26  & 	47.79  & 	23.74  & 	47.06  & 	71.22  & 	27.54  & 	15.96  & 	49.21 
\\
\hline

\rowcolor[HTML]{F2F2F2} 
\textbf{Exp Min}           & 0.00                & 0.00                                                                    & 0.00                                                                    & 0.00                                                                  & 0.00                                                                      & 0.00                                                                & 12.96                                                               & 6.41                                                                 \\
\rowcolor[HTML]{F2F2F2} 
\textbf{Exp Max}           & 100.00              & 100.00                                                                  & 100.00                                                                  & 100.00                                                                & 100.00                                                                    & 100.00                                                              & 33.84                                                               & 98.70                                         \\
\midrule
\end{tabular}
 }
 \label{tab:original}
 
\end{table}

\section{Experiments}

\textbf{Dimension Evaluation.}
For each image and instruction, we utilize official codes from various models for image editing. We calculate the \benchname scores following the methodology described in Sec.\ref{sec:eval}. The \benchname scores for original and diverse instructions are presented in Fig.~\ref{fig:score_radar}, Tab.\ref{tab:original}, and Tab.~\ref{tab:diverse}, respectively. Our observations reveal that no single model achieves the best performance across all evaluation dimensions.
Regarding low-level editing, InstructDiffusion~\cite{geng2023instructdiffusion} demonstrates superior results. It attains the highest scores in 4 out of 7 low-level editing evaluation dimensions when using original instructions, and 3 out of 7 when using diverse instructions.
For high-level editing, both MagicBrush~\cite{zhang2024magicbrush} and InstructAny2Pix~\cite{li2023instructany2pix} perform impressively. MagicBrush achieves the highest scores in 3 evaluation dimensions using original instructions, while InstructAny2Pix achieves the highest scores in 3 dimensions using diverse instructions.
In the deblurring dimensions, MGIE~\cite{fu2023guiding} stands out significantly. It surpasses the second-place model by 11.92 when using original instructions and by 11.37 when using diverse instructions.



\begin{table}[]
\caption{\benchname evaluation results per dimension using diverse instructions.}
\setlength{\tabcolsep}{2pt}
\resizebox{\linewidth}{!}{
\begin{tabular}{l|c|c|c|c|c|c|c|c}
\midrule
\multicolumn{9}{c}{\textbf{Low-level Editing}}           \\ 
\midrule
\textbf{Model}             & \textbf{Deblurring} & \textbf{\begin{tabular}[c]{@{}c@{}}Haze\\ Removal\end{tabular}}         & \textbf{\begin{tabular}[c]{@{}c@{}}Lowlight\\ Enhancement\end{tabular}} & \textbf{\begin{tabular}[c]{@{}c@{}}Noise\\ Removal\end{tabular}}      & \textbf{\begin{tabular}[c]{@{}c@{}}Rain\\ Removal\end{tabular}}           & \textbf{\begin{tabular}[c]{@{}c@{}}Shadow\\ Removal\end{tabular}}   & \textbf{\begin{tabular}[c]{@{}c@{}}Snow\\ Removal\end{tabular}}     & \textbf{\begin{tabular}[c]{@{}c@{}}Watermark\\ Removal\end{tabular}} \\
\midrule
\textbf{HIVE~\cite{zhang2023hive}}              & 44.41               & 54.09                                                                   & 42.78                                                                   & 25.51                                                                 & 58.59                                                                     & 36.69                                                               & 51.92                                                               & 57.88                                                                \\
\textbf{InstructDiffusion~\cite{geng2023instructdiffusion}} & 42.62               & 58.01                                                                   & {39.47}                                                          & \textbf{28.06}                                                        & {64.18}                                                            & 32.54                                                               & \textbf{57.30}                                                      & \textbf{85.14}                                                       \\
\textbf{InstructPix2Pix~\cite{brooks2023instructpix2pix}}   & 45.24               & 53.52                                                                   & \textbf{42.88}                                                          & 24.49                                                                 & 51.86                                                                     & 32.79                                                               & 52.67                                                               & 48.91                                                                \\
\textbf{MagicBrush~\cite{zhang2024magicbrush}}        & 45.96               & 55.11                                                          & 33.74                                                                   & 23.91                                                                 & 55.77                                                                     & \textbf{36.73}                                                      & {54.68}                                                      & 59.76                                                                \\
\textbf{MGIE~\cite{fu2023guiding}}              & \textbf{57.33}      & 51.61                                                                   & 32.96                                                                   & 23.49                                                                 & 58.27                                                                     & 34.07                                                               & 51.02                                                               & 59.64                                                                \\
\textbf{InstructEdit~\cite{wang2023instructedit}}      & 40.66               & \textbf{58.89}                                                          & 13.92                                                                   & 15.81                                                                 & \textbf{65.08}                                                            & {36.66}                                                      & 43.34                                                               & 83.68                                                                \\
\textbf{InstructAny2Pix~\cite{li2023instructany2pix}}   & 34.77               & 47.00                                                                   & 18.09                                                                   & 22.18                                                                 & 48.92                                                                     & 36.04                                                               & 43.13                                                               & 47.58                                                                \\
\textbf{HQ-Edit~\cite{hui2024hq}} & 34.11  & 	37.95  & 	36.76  & 	22.38  & 	37.60  & 	32.17  & 	38.45  & 	30.83 
\\
\hline
\rowcolor[HTML]{F2F2F2} 
\textbf{Exp Min}           & 6.32                & 3.67                                                                    & 0.60                                                                    & 0.03                                                                  & 7.22                                                                      & 1.46                                                                & 3.78                                                                & 2.58                                                                 \\
\rowcolor[HTML]{F2F2F2} 
\textbf{Exp Max}           & 88.17               & 92.69                                                                   & 90.34                                                                   & 79.29                                                                 & 97.03                                                                     & 86.27                                                               & 82.24                                                               & 96.39                                                                \\
\midrule
\multicolumn{9}{c}{\textbf{High-level Editing}}
\\
\midrule
\textbf{Model}             & \textbf{Counting}   & \textbf{\begin{tabular}[c]{@{}c@{}}Direction\\ Perception\end{tabular}} & \textbf{\begin{tabular}[c]{@{}c@{}}Object\\ Removal\end{tabular}}       & \textbf{\begin{tabular}[c]{@{}c@{}}Object\\ Replacement\end{tabular}} & \textbf{\begin{tabular}[c]{@{}c@{}}Background\\ Replacement\end{tabular}} & \textbf{\begin{tabular}[c]{@{}c@{}}Color\\ Alteration\end{tabular}} & \textbf{\begin{tabular}[c]{@{}c@{}}Style\\ Alteration\end{tabular}} & \textbf{\begin{tabular}[c]{@{}c@{}}Region\\ Accuracy\end{tabular}}   \\
\midrule
\textbf{HIVE~\cite{zhang2023hive}}              & 13.57               & 43.57                                                                   & 12.86                                                                   & \textbf{67.86}                                                        & \textbf{85.00}                                                            & 35.00                                                               & 23.08                                                               & 61.97                                                                \\
\textbf{InstructDiffusion~\cite{geng2023instructdiffusion}} & 21.43               & 47.86                                                                   & {22.14}                                                          & 47.14                                                                 & 64.29                                                                     & 48.57                                                               & 19.96                                                               & 65.92                                                                \\
\textbf{InstructPix2Pix~\cite{brooks2023instructpix2pix}}   & 18.57               & 47.86                                                                   & 7.14                                                                    & 47.14                                                                 & 65.71                                                                     & 43.57                                                               & 23.13                                                               & 61.32                                                                \\
\textbf{MagicBrush~\cite{zhang2024magicbrush}}        & \textbf{24.29}      & {45.71}                                                          & 12.14                                                                   & 62.14                                                                 & {83.57}                                                            & \textbf{54.29}                                                      & 23.08                                                               & 66.21                                                                \\
\textbf{MGIE~\cite{fu2023guiding}}              & 19.29               & 47.14                                                                   & {22.86}                                                          & 43.57                                                                 & 74.29                                                                     & 37.86                                                               & 23.36                                                               & 71.89                                                                \\
\textbf{InstructEdit~\cite{wang2023instructedit}}      & 11.76               & 46.04                                                                   & 3.60                                                                    & 4.41                                                                  & 51.80                                                                     & 3.62                                                                & 19.91                                                               & \textbf{77.08}                                                       \\
\textbf{InstructAny2Pix~\cite{li2023instructany2pix}}   & 22.79               & \textbf{51.80}                                                          & \textbf{43.88}                                                                   & 48.53                                                                 & 68.35                                                                     & 12.32                                                               & \textbf{25.93}                                                      & 52.61                                                                \\
\textbf{HQ-Edit~\cite{hui2024hq}} & 20.74  & 	51.47  & 	24.46  & 	50.00  & 	79.86  & 	26.09  & 	16.48  & 	48.29 
\\
\hline
\rowcolor[HTML]{F2F2F2} 
\textbf{Exp Min}           & 0.00                & 0.00                                                                    & 0.00                                                                    & 0.00                                                                  & 0.00                                                                      & 0.00                                                                & 10.62                                                               & 9.79
\\
\rowcolor[HTML]{F2F2F2} 
\textbf{Exp Max}           & 100.00              & 100.00                                                                  & 100.00                                                                  & 100.00                                                                & 100.00                                                                    & 100.00                                                              & 34.06                                                               & 98.68 \\
\midrule
\end{tabular}
}
\label{tab:diverse}
\end{table}

\begin{figure}
  \centering
  \includegraphics[width=1.0\columnwidth]{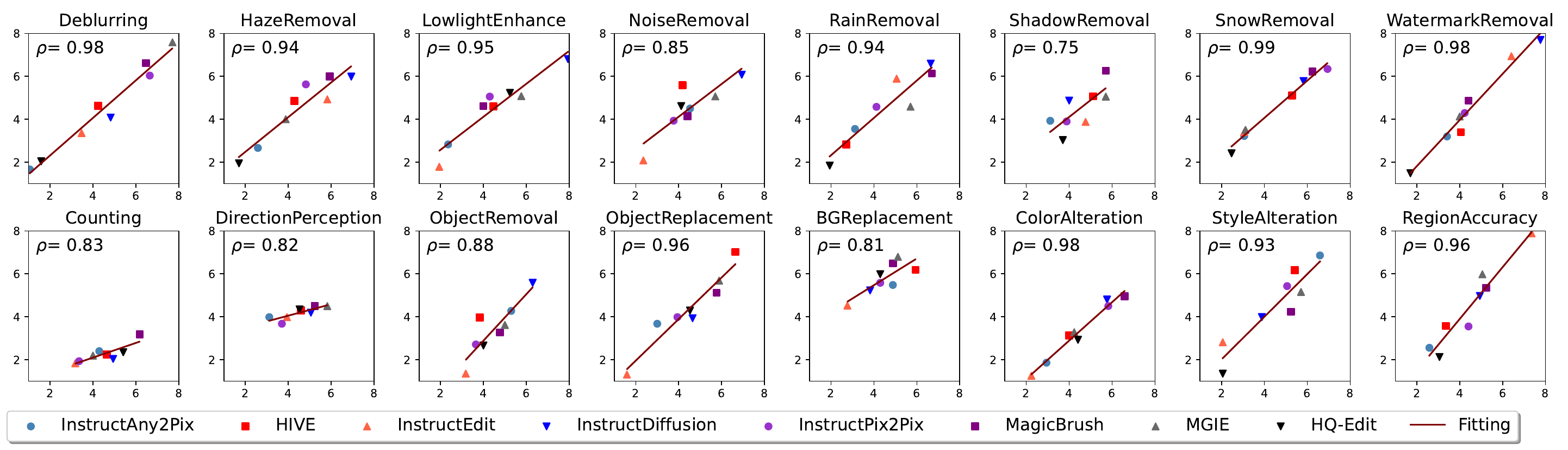}
  \caption{
    Alignment between \benchname rank scores (Y-axis) and human scores (X-axis).
  }
  \label{fig:ori_human}
\end{figure}

\textbf{Human Evaluation.}
We ranked different models based on their \benchname scores and computed \benchname rank scores using the methodology described in Sec.~\ref{sec:human}.
Given that both \benchname rank scores and human scores range from 1 to 8, a direct comparison can be made between them.
Therefore, we conducted correlation analyses and visually presented the results in Fig.~\ref{fig:ori_human}.
Significant positive correlations were observed between the \benchname rank score and the human score across all dimensions.
These findings offer strong evidence supporting the alignment between our proposed benchmark and human perception.


\section{Insights}

\textbf{The editing ability across different dimensions is not robust:}
Our observations indicate that no single model excels in all evaluation dimensions. This implies that different IIE models have varying strengths in terms of their editing abilities across different dimensions. Thus, it is crucial to acknowledge this limitation and focus on developing an IIE model that demonstrates consistent and competent performance across all dimensions. \textit{Future research efforts should prioritize the creation of a robust and versatile IIE model that can effectively handle a wide range of editing tasks across diverse dimensions.}

\begin{figure}
  \centering
  \includegraphics[width=1.0\columnwidth]{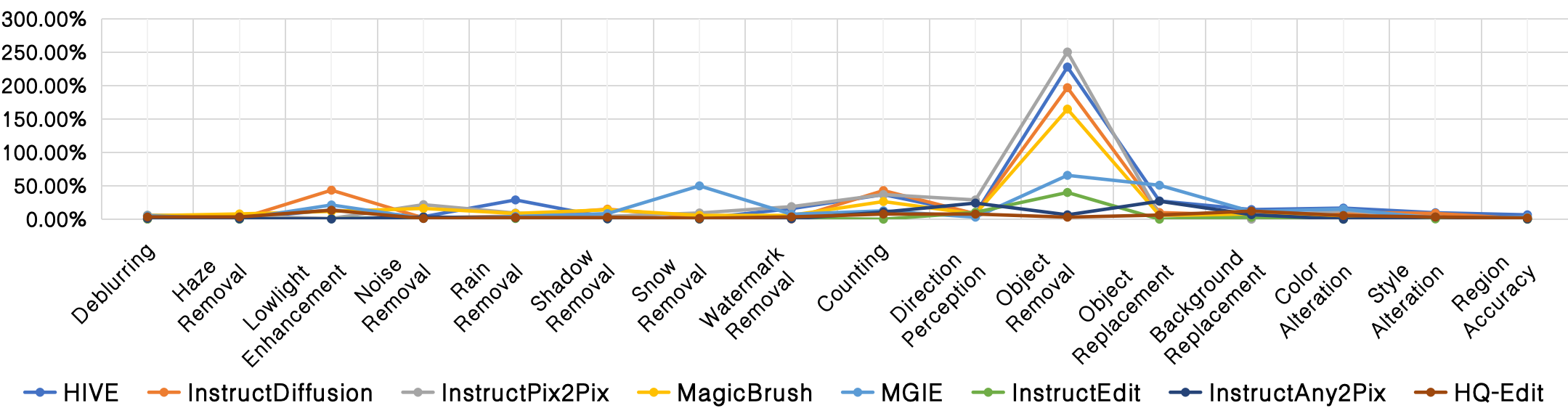}
  \caption{
    \benchname change rate using original instructions and diverse instructions.
  }
  \label{fig:score_change}
\end{figure}

\textbf{The editing ability of different instructions is not robust:}
To evaluate the robustness of editing models when provided with different instructions, we propose a metric called \benchname change rate. This metric is defined as follows:
\begin{equation}
    S^{i} = \frac{|S_o^{i}-S_d^{i}|}{\text{MIN}(S_o^{i}, S_d^{i})},
\end{equation}
where $S_o^i$ and $S_d^i$ represent the \benchname scores of the $i$-th evaluation dimension when using original and diverse instructions, respectively. The value of $S^i$ indicates the \benchname change rate for the $i$-th evaluation dimension. 
As illustrated in Fig.~\ref{fig:score_change}, when it comes to the object removal dimension, InstructPix2Pix\cite{brooks2023instructpix2pix}, HIVE~\cite{zhang2023hive}, InstructionDiffusion~\cite{geng2023instructdiffusion}, and MagicBrush~\cite{zhang2024magicbrush} exhibit significant fluctuations in their performance using different instructions.
On the other hand, the remaining models demonstrate relatively stable performance across different instructions.
One notable distinction between these two categories of models is that the latter employs LLM~\cite{achiam2023gpt,touvron2023llama} or MLLM~\cite{liu2023llava,liu2023improvedllava} to comprehend instructions, which enhances their resilience to variations in instructions.
\textit{Given the unpredictable and diverse nature of user editing instructions, it is crucial to develop an editing model that can effectively handle instructions with varying levels of complexity.
}
\begin{figure}
  \centering
  \includegraphics[width=1.0\columnwidth]{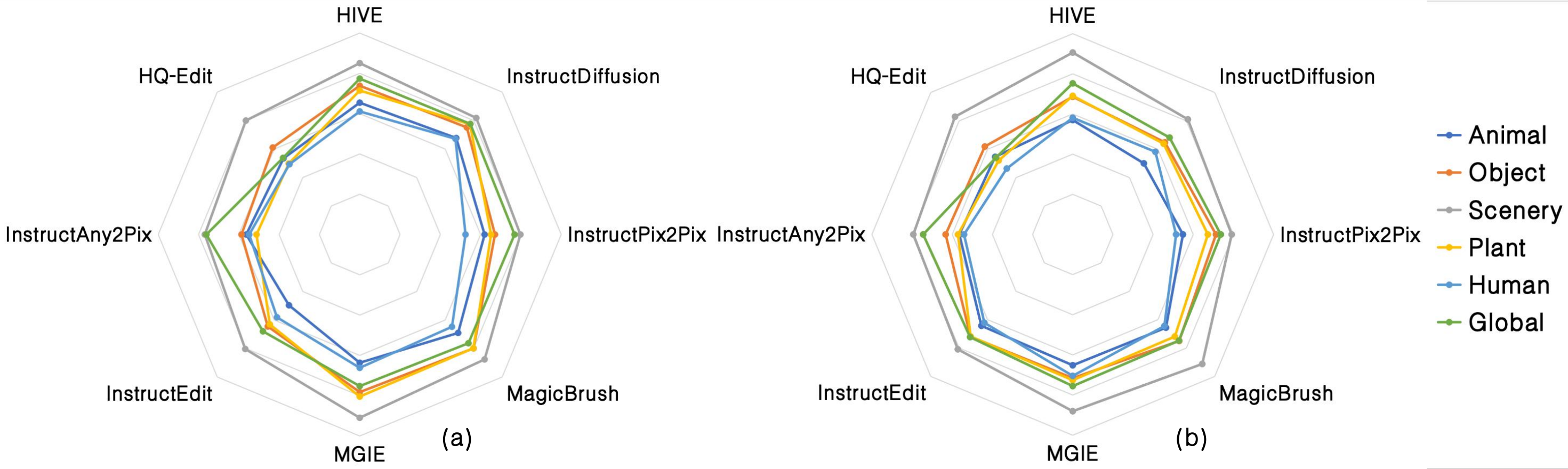}
  \caption{
    Comparison of radar charts for \benchname scores in different categories using (a) original instructions and (b) diverse instructions. The scores of all dimensions are normalized and averaged.
  }
  \label{fig:category_score}
\end{figure}

\textbf{The editing ability for different categories is not robust:}
As illustrated in Fig.~\ref{fig:category_score}, we have observed distinct variations in the performance of different categories. Notably, the "Scenery" and "Global" categories consistently demonstrate superior performance compared to the other categories across all the IIE models we evaluated. This discrepancy can be attributed to the inherent inclination of the "Scenery" and "Global" categories towards global editing, which diminishes the necessity for precise target object localization.
\textit{Given these findings, it is crucial to prioritize the simultaneous consideration of various editing content in future research endeavors.}


\section{Conclusions}
In this paper, we present \benchname, a comprehensive benchmark specifically designed for instruction-based image editing (IIE). Our benchmark includes a substantial dataset of over 2000+ images and more than 4000+ instructions, covering 16 distinct evaluation dimensions. To evaluate the effectiveness of \benchname, we conduct experiments using 8 open-source IIE models. Additionally, we complement these experiments with meticulous human evaluations to establish the correlation between \benchname scores and human perception.
Based on the observations derived from \benchname, we provide valuable insights and recommendations for advancing IIE models. 
We hope the proposed \benchname to serve as an indispensable asset, playing a pivotal role in fostering the advancement of IIE models and assessing their efficacy.

\bibliographystyle{plainnat} 
{
	\small
	\bibliography{egbib}

\begin{thebibliography}{100}
\providecommand{\natexlab}[1]{#1}
\providecommand{\url}[1]{\texttt{#1}}
\expandafter\ifx\csname urlstyle\endcsname\relax
  \providecommand{\doi}[1]{doi: #1}\else
  \providecommand{\doi}{doi: \begingroup \urlstyle{rm}\Url}\fi

\bibitem[Achiam et~al.(2023)Achiam, Adler, Agarwal, Ahmad, Akkaya, Aleman, Almeida, Altenschmidt, Altman, Anadkat, et~al.]{achiam2023gpt}
Josh Achiam, Steven Adler, Sandhini Agarwal, Lama Ahmad, Ilge Akkaya, Florencia~Leoni Aleman, Diogo Almeida, Janko Altenschmidt, Sam Altman, Shyamal Anadkat, et~al.
\newblock Gpt-4 technical report.
\newblock \emph{arXiv preprint arXiv:2303.08774}, 2023.

\bibitem[Alayrac et~al.(2022)Alayrac, Donahue, Luc, Miech, Barr, Hasson, Lenc, Mensch, Millican, Reynolds, et~al.]{alayrac2022flamingo}
Jean-Baptiste Alayrac, Jeff Donahue, Pauline Luc, Antoine Miech, Iain Barr, Yana Hasson, Karel Lenc, Arthur Mensch, Katherine Millican, Malcolm Reynolds, et~al.
\newblock Flamingo: a visual language model for few-shot learning.
\newblock \emph{Advances in neural information processing systems}, 35:\penalty0 23716--23736, 2022.

\bibitem[Ancuti et~al.(2019)Ancuti, Ancuti, Sbert, and Timofte]{ancuti2019dense}
Codruta~O Ancuti, Cosmin Ancuti, Mateu Sbert, and Radu Timofte.
\newblock Dense-haze: A benchmark for image dehazing with dense-haze and haze-free images.
\newblock In \emph{2019 IEEE international conference on image processing (ICIP)}, pages 1014--1018. IEEE, 2019.

\bibitem[Austin et~al.(2021)Austin, Johnson, Ho, Tarlow, and Van Den~Berg]{austin2021structured}
Jacob Austin, Daniel~D Johnson, Jonathan Ho, Daniel Tarlow, and Rianne Van Den~Berg.
\newblock Structured denoising diffusion models in discrete state-spaces.
\newblock \emph{Advances in Neural Information Processing Systems}, 34:\penalty0 17981--17993, 2021.

\bibitem[Avrahami et~al.(2022)Avrahami, Lischinski, and Fried]{avrahami2022blended}
Omri Avrahami, Dani Lischinski, and Ohad Fried.
\newblock Blended diffusion for text-driven editing of natural images.
\newblock In \emph{Proceedings of the IEEE/CVF Conference on Computer Vision and Pattern Recognition}, pages 18208--18218, 2022.

\bibitem[Bai et~al.(2023)Bai, Bai, Yang, Wang, Tan, Wang, Lin, Zhou, and Zhou]{bai2023qwen}
Jinze Bai, Shuai Bai, Shusheng Yang, Shijie Wang, Sinan Tan, Peng Wang, Junyang Lin, Chang Zhou, and Jingren Zhou.
\newblock Qwen-vl: A frontier large vision-language model with versatile abilities.
\newblock \emph{arXiv preprint arXiv:2308.12966}, 2023.

\bibitem[Basu et~al.(2023)Basu, Saberi, Bhardwaj, Chegini, Massiceti, Sanjabi, Hu, and Feizi]{basu2023editval}
Samyadeep Basu, Mehrdad Saberi, Shweta Bhardwaj, Atoosa~Malemir Chegini, Daniela Massiceti, Maziar Sanjabi, Shell~Xu Hu, and Soheil Feizi.
\newblock Editval: Benchmarking diffusion based text-guided image editing methods.
\newblock \emph{arXiv preprint arXiv:2310.02426}, 2023.

\bibitem[Betker et~al.(2023)Betker, Goh, Jing, Brooks, Wang, Li, Ouyang, Zhuang, Lee, Guo, et~al.]{betker2023improving}
James Betker, Gabriel Goh, Li~Jing, Tim Brooks, Jianfeng Wang, Linjie Li, Long Ouyang, Juntang Zhuang, Joyce Lee, Yufei Guo, et~al.
\newblock Improving image generation with better captions.
\newblock \emph{Computer Science. https://cdn. openai. com/papers/dall-e-3. pdf}, 2\penalty0 (3):\penalty0 8, 2023.

\bibitem[Bigham et~al.(2010)Bigham, Jayant, Ji, Little, Miller, Miller, Miller, Tatarowicz, White, White, et~al.]{bigham2010vizwiz}
Jeffrey~P Bigham, Chandrika Jayant, Hanjie Ji, Greg Little, Andrew Miller, Robert~C Miller, Robin Miller, Aubrey Tatarowicz, Brandyn White, Samual White, et~al.
\newblock Vizwiz: nearly real-time answers to visual questions.
\newblock In \emph{Proceedings of the 23nd annual ACM symposium on User interface software and technology}, pages 333--342, 2010.

\bibitem[Brooks et~al.(2023)Brooks, Holynski, and Efros]{brooks2023instructpix2pix}
Tim Brooks, Aleksander Holynski, and Alexei~A Efros.
\newblock Instructpix2pix: Learning to follow image editing instructions.
\newblock In \emph{Proceedings of the IEEE/CVF Conference on Computer Vision and Pattern Recognition}, pages 18392--18402, 2023.

\bibitem[Brown et~al.(2020)Brown, Mann, Ryder, Subbiah, Kaplan, Dhariwal, Neelakantan, Shyam, Sastry, Askell, et~al.]{brown2020language}
Tom Brown, Benjamin Mann, Nick Ryder, Melanie Subbiah, Jared~D Kaplan, Prafulla Dhariwal, Arvind Neelakantan, Pranav Shyam, Girish Sastry, Amanda Askell, et~al.
\newblock Language models are few-shot learners.
\newblock \emph{Advances in neural information processing systems}, 33:\penalty0 1877--1901, 2020.

\bibitem[Chen et~al.(2023{\natexlab{a}})Chen, Gu, Liu, Magid, Dong, Wang, Pfister, and Zhu]{chen2023masked}
Haoyu Chen, Jinjin Gu, Yihao Liu, Salma~Abdel Magid, Chao Dong, Qiong Wang, Hanspeter Pfister, and Lei Zhu.
\newblock Masked image training for generalizable deep image denoising.
\newblock In \emph{Proceedings of the IEEE/CVF Conference on Computer Vision and Pattern Recognition}, pages 1692--1703, 2023{\natexlab{a}}.

\bibitem[Chen et~al.(2021)Chen, Fang, Hsieh, Tsai, Chen, Ding, Kuo, et~al.]{chen2021all}
Wei-Ting Chen, Hao-Yu Fang, Cheng-Lin Hsieh, Cheng-Che Tsai, I~Chen, Jian-Jiun Ding, Sy-Yen Kuo, et~al.
\newblock All snow removed: Single image desnowing algorithm using hierarchical dual-tree complex wavelet representation and contradict channel loss.
\newblock In \emph{Proceedings of the IEEE/CVF International Conference on Computer Vision}, pages 4196--4205, 2021.

\bibitem[Chen et~al.(2023{\natexlab{b}})Chen, Li, Li, and Pan]{chen2023learning}
Xiang Chen, Hao Li, Mingqiang Li, and Jinshan Pan.
\newblock Learning a sparse transformer network for effective image deraining.
\newblock In \emph{Proceedings of the IEEE/CVF Conference on Computer Vision and Pattern Recognition}, pages 5896--5905, 2023{\natexlab{b}}.

\bibitem[Chen et~al.(2018)Chen, Lai, and Liu]{chen2018cartoongan}
Yang Chen, Yu-Kun Lai, and Yong-Jin Liu.
\newblock Cartoongan: Generative adversarial networks for photo cartoonization.
\newblock In \emph{Proceedings of the IEEE conference on computer vision and pattern recognition}, pages 9465--9474, 2018.

\bibitem[Chu et~al.(2024)Chu, Qiao, Zhang, Xu, Wei, Yang, Sun, Hu, Lin, Zhang, et~al.]{chu2024mobilevlm}
Xiangxiang Chu, Limeng Qiao, Xinyu Zhang, Shuang Xu, Fei Wei, Yang Yang, Xiaofei Sun, Yiming Hu, Xinyang Lin, Bo~Zhang, et~al.
\newblock Mobilevlm v2: Faster and stronger baseline for vision language model.
\newblock \emph{arXiv preprint arXiv:2402.03766}, 2024.

\bibitem[Dai et~al.(2024)Dai, Li, Li, Tiong, Zhao, Wang, Li, Fung, and Hoi]{dai2024instructblip}
Wenliang Dai, Junnan Li, Dongxu Li, Anthony Meng~Huat Tiong, Junqi Zhao, Weisheng Wang, Boyang Li, Pascale~N Fung, and Steven Hoi.
\newblock Instructblip: Towards general-purpose vision-language models with instruction tuning.
\newblock \emph{Advances in Neural Information Processing Systems}, 36, 2024.

\bibitem[Dockhorn et~al.(2022)Dockhorn, Vahdat, and Kreis]{dockhorn2022genie}
Tim Dockhorn, Arash Vahdat, and Karsten Kreis.
\newblock Genie: Higher-order denoising diffusion solvers.
\newblock \emph{Advances in Neural Information Processing Systems}, 35:\penalty0 30150--30166, 2022.

\bibitem[Dong et~al.(2023)Dong, Xue, Duan, and Han]{dong2023prompt}
Wenkai Dong, Song Xue, Xiaoyue Duan, and Shumin Han.
\newblock Prompt tuning inversion for text-driven image editing using diffusion models.
\newblock In \emph{Proceedings of the IEEE/CVF International Conference on Computer Vision}, pages 7430--7440, 2023.

\bibitem[Dong et~al.(2024)Dong, Zhang, Zang, Cao, Wang, Ouyang, Wei, Zhang, Duan, Cao, et~al.]{dong2024internlm}
Xiaoyi Dong, Pan Zhang, Yuhang Zang, Yuhang Cao, Bin Wang, Linke Ouyang, Xilin Wei, Songyang Zhang, Haodong Duan, Maosong Cao, et~al.
\newblock Internlm-xcomposer2: Mastering free-form text-image composition and comprehension in vision-language large model.
\newblock \emph{arXiv preprint arXiv:2401.16420}, 2024.

\bibitem[Fei et~al.(2024{\natexlab{a}})Fei, Wu, Ji, Zhang, Zhang, Lee, and Hsu]{fei2024video}
Hao Fei, Shengqiong Wu, Wei Ji, Hanwang Zhang, Meishan Zhang, Mong-Li Lee, and Wynne Hsu.
\newblock Video-of-thought: Step-by-step video reasoning from perception to cognition.
\newblock In \emph{Forty-first International Conference on Machine Learning}, 2024{\natexlab{a}}.

\bibitem[Fei et~al.(2024{\natexlab{b}})Fei, Wu, Zhang, Chua, and Yan]{fei2024vitron}
Hao Fei, Shengqiong Wu, Hanwang Zhang, Tat-Seng Chua, and Shuicheng Yan.
\newblock Vitron: A unified pixel-level vision llm for understanding, generating, segmenting, editing, 2024{\natexlab{b}}.

\bibitem[Fei et~al.(2024{\natexlab{c}})Fei, Wu, Zhang, Zhang, Chua, and Yan]{fei2024enhancing}
Hao Fei, Shengqiong Wu, Meishan Zhang, Min Zhang, Tat-Seng Chua, and Shuicheng Yan.
\newblock Enhancing video-language representations with structural spatio-temporal alignment.
\newblock \emph{IEEE Transactions on Pattern Analysis and Machine Intelligence}, 2024{\natexlab{c}}.

\bibitem[Fu et~al.(2024)Fu, Hu, Du, Wang, Yang, and Gan]{fu2023guiding}
Tsu-Jui Fu, Wenze Hu, Xianzhi Du, William~Yang Wang, Yinfei Yang, and Zhe Gan.
\newblock Guiding instruction-based image editing via multimodal large language models.
\newblock \emph{International Conference on Learning Representations}, 2024.

\bibitem[Gao et~al.(2024)Gao, Zhang, Liu, Qiu, Huang, Lin, Zhao, Geng, Lin, Jin, et~al.]{gao2024sphinx}
Peng Gao, Renrui Zhang, Chris Liu, Longtian Qiu, Siyuan Huang, Weifeng Lin, Shitian Zhao, Shijie Geng, Ziyi Lin, Peng Jin, et~al.
\newblock Sphinx-x: Scaling data and parameters for a family of multi-modal large language models.
\newblock \emph{arXiv preprint arXiv:2402.05935}, 2024.

\bibitem[Geng et~al.(2023)Geng, Yang, Hang, Li, Gu, Zhang, Bao, Zhang, Hu, Chen, et~al.]{geng2023instructdiffusion}
Zigang Geng, Binxin Yang, Tiankai Hang, Chen Li, Shuyang Gu, Ting Zhang, Jianmin Bao, Zheng Zhang, Han Hu, Dong Chen, et~al.
\newblock Instructdiffusion: A generalist modeling interface for vision tasks.
\newblock \emph{arXiv preprint arXiv:2309.03895}, 2023.

\bibitem[Goodfellow et~al.(2014)Goodfellow, Pouget-Abadie, Mirza, Xu, Warde-Farley, Ozair, Courville, and Bengio]{goodfellow2014generative}
Ian Goodfellow, Jean Pouget-Abadie, Mehdi Mirza, Bing Xu, David Warde-Farley, Sherjil Ozair, Aaron Courville, and Yoshua Bengio.
\newblock Generative adversarial nets.
\newblock \emph{Advances in neural information processing systems}, 27, 2014.

\bibitem[Goodfellow et~al.(2020)Goodfellow, Pouget-Abadie, Mirza, Xu, Warde-Farley, Ozair, Courville, and Bengio]{goodfellow2020generative}
Ian Goodfellow, Jean Pouget-Abadie, Mehdi Mirza, Bing Xu, David Warde-Farley, Sherjil Ozair, Aaron Courville, and Yoshua Bengio.
\newblock Generative adversarial networks.
\newblock \emph{Communications of the ACM}, 63\penalty0 (11):\penalty0 139--144, 2020.

\bibitem[Guo et~al.(2023{\natexlab{a}})Guo, Wang, Yang, Huang, Wang, Pfister, and Wen]{guo2023shadowdiffusion}
Lanqing Guo, Chong Wang, Wenhan Yang, Siyu Huang, Yufei Wang, Hanspeter Pfister, and Bihan Wen.
\newblock Shadowdiffusion: When degradation prior meets diffusion model for shadow removal.
\newblock In \emph{Proceedings of the IEEE/CVF Conference on Computer Vision and Pattern Recognition}, pages 14049--14058, 2023{\natexlab{a}}.

\bibitem[Guo et~al.(2023{\natexlab{b}})Guo, Xiao, Chang, Deng, and Yan]{guo2023sky}
Yun Guo, Xueyao Xiao, Yi~Chang, Shumin Deng, and Luxin Yan.
\newblock From sky to the ground: A large-scale benchmark and simple baseline towards real rain removal.
\newblock In \emph{Proceedings of the IEEE/CVF International Conference on Computer Vision}, pages 12097--12107, 2023{\natexlab{b}}.

\bibitem[Hertz et~al.(2022)Hertz, Mokady, Tenenbaum, Aberman, Pritch, and Cohen-Or]{hertz2022prompt}
Amir Hertz, Ron Mokady, Jay Tenenbaum, Kfir Aberman, Yael Pritch, and Daniel Cohen-Or.
\newblock Prompt-to-prompt image editing with cross attention control.
\newblock \emph{arXiv preprint arXiv:2208.01626}, 2022.

\bibitem[Ho et~al.(2020)Ho, Jain, and Abbeel]{ho2020denoising}
Jonathan Ho, Ajay Jain, and Pieter Abbeel.
\newblock Denoising diffusion probabilistic models.
\newblock \emph{Advances in neural information processing systems}, 33:\penalty0 6840--6851, 2020.

\bibitem[Hu et~al.(2024)Hu, Xu, Ye, Yan, Zhang, Zhang, Li, Zhang, Jin, Huang, et~al.]{hu2024mplug}
Anwen Hu, Haiyang Xu, Jiabo Ye, Ming Yan, Liang Zhang, Bo~Zhang, Chen Li, Ji~Zhang, Qin Jin, Fei Huang, et~al.
\newblock mplug-docowl 1.5: Unified structure learning for ocr-free document understanding.
\newblock \emph{arXiv preprint arXiv:2403.12895}, 2024.

\bibitem[Huang et~al.(2024{\natexlab{a}})Huang, Huang, Liu, Yan, Lv, Liu, Xiong, Zhang, Chen, and Cao]{huang2024diffusion}
Yi~Huang, Jiancheng Huang, Yifan Liu, Mingfu Yan, Jiaxi Lv, Jianzhuang Liu, Wei Xiong, He~Zhang, Shifeng Chen, and Liangliang Cao.
\newblock Diffusion model-based image editing: A survey.
\newblock \emph{arXiv preprint arXiv:2402.17525}, 2024{\natexlab{a}}.

\bibitem[Huang et~al.(2024{\natexlab{b}})Huang, Xie, Wang, Yuan, Cun, Ge, Zhou, Dong, Huang, Zhang, et~al.]{huang2023smartedit}
Yuzhou Huang, Liangbin Xie, Xintao Wang, Ziyang Yuan, Xiaodong Cun, Yixiao Ge, Jiantao Zhou, Chao Dong, Rui Huang, Ruimao Zhang, et~al.
\newblock Smartedit: Exploring complex instruction-based image editing with multimodal large language models.
\newblock \emph{Proceedings of the IEEE/CVF Conference on Computer Vision and Pattern Recognition}, 2024{\natexlab{b}}.

\bibitem[Hudson and Manning(2019)]{hudson2019gqa}
Drew~A Hudson and Christopher~D Manning.
\newblock Gqa: A new dataset for real-world visual reasoning and compositional question answering.
\newblock In \emph{Proceedings of the IEEE/CVF conference on computer vision and pattern recognition}, pages 6700--6709, 2019.

\bibitem[Hui et~al.(2024)Hui, Yang, Zhao, Shi, Wang, Wang, Zhou, and Xie]{hui2024hq}
Mude Hui, Siwei Yang, Bingchen Zhao, Yichun Shi, Heng Wang, Peng Wang, Yuyin Zhou, and Cihang Xie.
\newblock Hq-edit: A high-quality dataset for instruction-based image editing.
\newblock \emph{arXiv preprint arXiv:2404.09990}, 2024.

\bibitem[Ji et~al.(2022)Ji, Ma, Sun, Zhou, Wu, and Ji]{ji2022knowing}
Jiayi Ji, Yiwei Ma, Xiaoshuai Sun, Yiyi Zhou, Yongjian Wu, and Rongrong Ji.
\newblock Knowing what to learn: a metric-oriented focal mechanism for image captioning.
\newblock \emph{IEEE Transactions on Image Processing}, 31:\penalty0 4321--4335, 2022.

\bibitem[Karras et~al.(2019)Karras, Laine, and Aila]{karras2019style}
Tero Karras, Samuli Laine, and Timo Aila.
\newblock A style-based generator architecture for generative adversarial networks.
\newblock In \emph{Proceedings of the IEEE/CVF conference on computer vision and pattern recognition}, pages 4401--4410, 2019.

\bibitem[Karras et~al.(2020{\natexlab{a}})Karras, Aittala, Hellsten, Laine, Lehtinen, and Aila]{karras2020training}
Tero Karras, Miika Aittala, Janne Hellsten, Samuli Laine, Jaakko Lehtinen, and Timo Aila.
\newblock Training generative adversarial networks with limited data.
\newblock \emph{Advances in neural information processing systems}, 33:\penalty0 12104--12114, 2020{\natexlab{a}}.

\bibitem[Karras et~al.(2020{\natexlab{b}})Karras, Laine, Aittala, Hellsten, Lehtinen, and Aila]{karras2020analyzing}
Tero Karras, Samuli Laine, Miika Aittala, Janne Hellsten, Jaakko Lehtinen, and Timo Aila.
\newblock Analyzing and improving the image quality of stylegan.
\newblock In \emph{Proceedings of the IEEE/CVF conference on computer vision and pattern recognition}, pages 8110--8119, 2020{\natexlab{b}}.

\bibitem[Kawar et~al.(2022)Kawar, Elad, Ermon, and Song]{kawar2022denoising}
Bahjat Kawar, Michael Elad, Stefano Ermon, and Jiaming Song.
\newblock Denoising diffusion restoration models.
\newblock \emph{Advances in Neural Information Processing Systems}, 35:\penalty0 23593--23606, 2022.

\bibitem[Kawar et~al.(2023)Kawar, Zada, Lang, Tov, Chang, Dekel, Mosseri, and Irani]{kawar2023imagic}
Bahjat Kawar, Shiran Zada, Oran Lang, Omer Tov, Huiwen Chang, Tali Dekel, Inbar Mosseri, and Michal Irani.
\newblock Imagic: Text-based real image editing with diffusion models.
\newblock In \emph{Proceedings of the IEEE/CVF Conference on Computer Vision and Pattern Recognition}, pages 6007--6017, 2023.

\bibitem[Kong et~al.(2022)Kong, Liu, Gu, Qiao, and Dong]{kong2022reflash}
Xiangtao Kong, Xina Liu, Jinjin Gu, Yu~Qiao, and Chao Dong.
\newblock Reflash dropout in image super-resolution.
\newblock In \emph{Proceedings of the IEEE/CVF Conference on Computer Vision and Pattern Recognition}, pages 6002--6012, 2022.

\bibitem[Korhonen and You(2012)]{korhonen2012peak}
Jari Korhonen and Junyong You.
\newblock Peak signal-to-noise ratio revisited: Is simple beautiful?
\newblock In \emph{2012 Fourth International Workshop on Quality of Multimedia Experience}, pages 37--38. IEEE, 2012.

\bibitem[Kulikov et~al.(2023)Kulikov, Yadin, Kleiner, and Michaeli]{kulikov2023sinddm}
Vladimir Kulikov, Shahar Yadin, Matan Kleiner, and Tomer Michaeli.
\newblock Sinddm: A single image denoising diffusion model.
\newblock In \emph{International Conference on Machine Learning (ICML)}, pages 17920--17930. PMLR, 2023.

\bibitem[Li et~al.(2023{\natexlab{a}})Li, Wang, Wang, Ge, Ge, and Shan]{li2023seed}
Bohao Li, Rui Wang, Guangzhi Wang, Yuying Ge, Yixiao Ge, and Ying Shan.
\newblock Seed-bench: Benchmarking multimodal llms with generative comprehension.
\newblock \emph{arXiv preprint arXiv:2307.16125}, 2023{\natexlab{a}}.

\bibitem[Li et~al.(2023{\natexlab{b}})Li, Li, Savarese, and Hoi]{li2023blip}
Junnan Li, Dongxu Li, Silvio Savarese, and Steven Hoi.
\newblock Blip-2: Bootstrapping language-image pre-training with frozen image encoders and large language models.
\newblock In \emph{International conference on machine learning}, pages 19730--19742. PMLR, 2023{\natexlab{b}}.

\bibitem[Li et~al.(2023{\natexlab{c}})Li, Wang, He, Li, Wang, Liu, Wang, Xu, Chen, Luo, et~al.]{li2023mvbench}
Kunchang Li, Yali Wang, Yinan He, Yizhuo Li, Yi~Wang, Yi~Liu, Zun Wang, Jilan Xu, Guo Chen, Ping Luo, et~al.
\newblock Mvbench: A comprehensive multi-modal video understanding benchmark.
\newblock \emph{arXiv preprint arXiv:2311.17005}, 2023{\natexlab{c}}.

\bibitem[Li et~al.(2023{\natexlab{d}})Li, Singh, and Grover]{li2023instructany2pix}
Shufan Li, Harkanwar Singh, and Aditya Grover.
\newblock Instructany2pix: Flexible visual editing via multimodal instruction following.
\newblock \emph{arXiv preprint arXiv:2312.06738}, 2023{\natexlab{d}}.

\bibitem[Li et~al.(2023{\natexlab{e}})Li, Du, Zhou, Wang, Zhao, and Wen]{li2023evaluating}
Yifan Li, Yifan Du, Kun Zhou, Jinpeng Wang, Wayne~Xin Zhao, and Ji-Rong Wen.
\newblock Evaluating object hallucination in large vision-language models.
\newblock \emph{arXiv preprint arXiv:2305.10355}, 2023{\natexlab{e}}.

\bibitem[Lin et~al.(2014)Lin, Maire, Belongie, Hays, Perona, Ramanan, Doll{\'a}r, and Zitnick]{lin2014microsoft}
Tsung-Yi Lin, Michael Maire, Serge Belongie, James Hays, Pietro Perona, Deva Ramanan, Piotr Doll{\'a}r, and C~Lawrence Zitnick.
\newblock Microsoft coco: Common objects in context.
\newblock In \emph{Computer Vision--ECCV 2014: 13th European Conference, Zurich, Switzerland, September 6-12, 2014, Proceedings, Part V 13}, pages 740--755. Springer, 2014.

\bibitem[Liu et~al.(2023{\natexlab{a}})Liu, Li, Li, and Lee]{liu2023improvedllava}
Haotian Liu, Chunyuan Li, Yuheng Li, and Yong~Jae Lee.
\newblock Improved baselines with visual instruction tuning, 2023{\natexlab{a}}.

\bibitem[Liu et~al.(2023{\natexlab{b}})Liu, Li, Wu, and Lee]{liu2023llava}
Haotian Liu, Chunyuan Li, Qingyang Wu, and Yong~Jae Lee.
\newblock Visual instruction tuning, 2023{\natexlab{b}}.

\bibitem[Liu et~al.(2021{\natexlab{a}})Liu, Zhu, and Bai]{Liu_2021_WACV}
Yang Liu, Zhen Zhu, and Xiang Bai.
\newblock Wdnet: Watermark-decomposition network for visible watermark removal.
\newblock In \emph{2021 {IEEE/CVF} Winter Conference on Applications of Computer Vision (WACV)}. {IEEE}, 2021{\natexlab{a}}.

\bibitem[Liu et~al.(2021{\natexlab{b}})Liu, Zhu, Pei, Fu, Qin, Zhang, Wan, and Feng]{liu2021synthetic}
Ye~Liu, Lei Zhu, Shunda Pei, Huazhu Fu, Jing Qin, Qing Zhang, Liang Wan, and Wei Feng.
\newblock From synthetic to real: Image dehazing collaborating with unlabeled real data.
\newblock In \emph{Proceedings of the 29th ACM international conference on multimedia}, pages 50--58, 2021{\natexlab{b}}.

\bibitem[Lu et~al.(2022)Lu, Mishra, Xia, Qiu, Chang, Zhu, Tafjord, Clark, and Kalyan]{lu2022learn}
Pan Lu, Swaroop Mishra, Tanglin Xia, Liang Qiu, Kai-Wei Chang, Song-Chun Zhu, Oyvind Tafjord, Peter Clark, and Ashwin Kalyan.
\newblock Learn to explain: Multimodal reasoning via thought chains for science question answering.
\newblock \emph{Advances in Neural Information Processing Systems}, 35:\penalty0 2507--2521, 2022.

\bibitem[Ma et~al.(2022)Ma, Xu, Sun, Yan, Zhang, and Ji]{ma2022x}
Yiwei Ma, Guohai Xu, Xiaoshuai Sun, Ming Yan, Ji~Zhang, and Rongrong Ji.
\newblock X-clip: End-to-end multi-grained contrastive learning for video-text retrieval.
\newblock In \emph{Proceedings of the 30th ACM International Conference on Multimedia}, pages 638--647, 2022.

\bibitem[Ma et~al.(2023)Ma, Ji, Sun, Zhou, and Ji]{ma2023towards}
Yiwei Ma, Jiayi Ji, Xiaoshuai Sun, Yiyi Zhou, and Rongrong Ji.
\newblock Towards local visual modeling for image captioning.
\newblock \emph{Pattern Recognition}, 138:\penalty0 109420, 2023.

\bibitem[Ma et~al.(2024)Ma, Wang, Sun, Lin, Zhou, Ji, and Ji]{ma2024inf}
Yiwei Ma, Zhibin Wang, Xiaoshuai Sun, Weihuang Lin, Qiang Zhou, Jiayi Ji, and Rongrong Ji.
\newblock Inf-llava: Dual-perspective perception for high-resolution multimodal large language model.
\newblock \emph{arXiv preprint arXiv:2407.16198}, 2024.

\bibitem[Mao et~al.(2017)Mao, Li, Xie, Lau, Wang, and Paul~Smolley]{mao2017least}
Xudong Mao, Qing Li, Haoran Xie, Raymond~YK Lau, Zhen Wang, and Stephen Paul~Smolley.
\newblock Least squares generative adversarial networks.
\newblock In \emph{Proceedings of the IEEE international conference on computer vision}, pages 2794--2802, 2017.

\bibitem[Marino et~al.(2019)Marino, Rastegari, Farhadi, and Mottaghi]{marino2019ok}
Kenneth Marino, Mohammad Rastegari, Ali Farhadi, and Roozbeh Mottaghi.
\newblock Ok-vqa: A visual question answering benchmark requiring external knowledge.
\newblock In \emph{Proceedings of the IEEE/cvf conference on computer vision and pattern recognition}, pages 3195--3204, 2019.

\bibitem[Martin et~al.(2001)Martin, Fowlkes, Tal, and Malik]{MartinFTM01}
D.~Martin, C.~Fowlkes, D.~Tal, and J.~Malik.
\newblock A database of human segmented natural images and its application to evaluating segmentation algorithms and measuring ecological statistics.
\newblock In \emph{Proc. 8th Int'l Conf. Computer Vision}, volume~2, pages 416--423, July 2001.

\bibitem[Nah et~al.(2017)Nah, Kim, and Lee]{Nah_2017_CVPR}
Seungjun Nah, Tae~Hyun Kim, and Kyoung~Mu Lee.
\newblock Deep multi-scale convolutional neural network for dynamic scene deblurring.
\newblock In \emph{The IEEE Conference on Computer Vision and Pattern Recognition (CVPR)}, July 2017.

\bibitem[Nichol and Dhariwal(2021)]{nichol2021improved}
Alexander~Quinn Nichol and Prafulla Dhariwal.
\newblock Improved denoising diffusion probabilistic models.
\newblock In \emph{International conference on machine learning}, pages 8162--8171. PMLR, 2021.

\bibitem[Qu et~al.(2017)Qu, Tian, He, Tang, and Lau]{qu2017deshadownet}
Liangqiong Qu, Jiandong Tian, Shengfeng He, Yandong Tang, and Rynson~WH Lau.
\newblock Deshadownet: A multi-context embedding deep network for shadow removal.
\newblock In \emph{Proceedings of the IEEE conference on computer vision and pattern recognition}, pages 4067--4075, 2017.

\bibitem[Radford et~al.(2021)Radford, Kim, Hallacy, Ramesh, Goh, Agarwal, Sastry, Askell, Mishkin, Clark, et~al.]{radford2021learning}
Alec Radford, Jong~Wook Kim, Chris Hallacy, Aditya Ramesh, Gabriel Goh, Sandhini Agarwal, Girish Sastry, Amanda Askell, Pamela Mishkin, Jack Clark, et~al.
\newblock Learning transferable visual models from natural language supervision.
\newblock In \emph{International conference on machine learning (ICML)}, pages 8748--8763. PMLR, 2021.

\bibitem[Ramesh et~al.(2021)Ramesh, Pavlov, Goh, Gray, Voss, Radford, Chen, and Sutskever]{ramesh2021zero}
Aditya Ramesh, Mikhail Pavlov, Gabriel Goh, Scott Gray, Chelsea Voss, Alec Radford, Mark Chen, and Ilya Sutskever.
\newblock Zero-shot text-to-image generation.
\newblock In \emph{International conference on machine learning}, pages 8821--8831. Pmlr, 2021.

\bibitem[Ramesh et~al.(2022)Ramesh, Dhariwal, Nichol, Chu, and Chen]{ramesh2022hierarchical}
Aditya Ramesh, Prafulla Dhariwal, Alex Nichol, Casey Chu, and Mark Chen.
\newblock Hierarchical text-conditional image generation with clip latents.
\newblock \emph{arXiv preprint arXiv:2204.06125}, 1\penalty0 (2):\penalty0 3, 2022.

\bibitem[Reid et~al.(2024)Reid, Savinov, Teplyashin, Lepikhin, Lillicrap, Alayrac, Soricut, Lazaridou, Firat, Schrittwieser, et~al.]{reid2024gemini}
Machel Reid, Nikolay Savinov, Denis Teplyashin, Dmitry Lepikhin, Timothy Lillicrap, Jean-baptiste Alayrac, Radu Soricut, Angeliki Lazaridou, Orhan Firat, Julian Schrittwieser, et~al.
\newblock Gemini 1.5: Unlocking multimodal understanding across millions of tokens of context.
\newblock \emph{arXiv preprint arXiv:2403.05530}, 2024.

\bibitem[Rombach et~al.(2022)Rombach, Blattmann, Lorenz, Esser, and Ommer]{rombach2022high}
Robin Rombach, Andreas Blattmann, Dominik Lorenz, Patrick Esser, and Bj{\"o}rn Ommer.
\newblock High-resolution image synthesis with latent diffusion models.
\newblock In \emph{Proceedings of the IEEE/CVF conference on computer vision and pattern recognition}, pages 10684--10695, 2022.

\bibitem[Saharia et~al.(2022)Saharia, Chan, Saxena, Li, Whang, Denton, Ghasemipour, Gontijo~Lopes, Karagol~Ayan, Salimans, et~al.]{saharia2022photorealistic}
Chitwan Saharia, William Chan, Saurabh Saxena, Lala Li, Jay Whang, Emily~L Denton, Kamyar Ghasemipour, Raphael Gontijo~Lopes, Burcu Karagol~Ayan, Tim Salimans, et~al.
\newblock Photorealistic text-to-image diffusion models with deep language understanding.
\newblock \emph{Advances in neural information processing systems}, 35:\penalty0 36479--36494, 2022.

\bibitem[Sanghvi et~al.(2023)Sanghvi, Mao, and Chan]{sanghvi2023structured}
Yash Sanghvi, Zhiyuan Mao, and Stanley~H Chan.
\newblock Structured kernel estimation for photon-limited deconvolution.
\newblock In \emph{Proceedings of the IEEE/CVF Conference on Computer Vision and Pattern Recognition}, pages 9863--9872, 2023.

\bibitem[Saund et~al.(2003)Saund, Fleet, Larner, and Mahoney]{saund2003perceptually}
Eric Saund, David Fleet, Daniel Larner, and James Mahoney.
\newblock Perceptually-supported image editing of text and graphics.
\newblock In \emph{Proceedings of the 16th annual ACM symposium on User interface software and technology}, pages 183--192, 2003.

\bibitem[Shen et~al.(2019)Shen, Wang, Lu, Shen, Ling, Xu, and Shao]{shen2019human}
Ziyi Shen, Wenguan Wang, Xiankai Lu, Jianbing Shen, Haibin Ling, Tingfa Xu, and Ling Shao.
\newblock Human-aware motion deblurring.
\newblock In \emph{Proceedings of the IEEE/CVF International Conference on Computer Vision}, pages 5572--5581, 2019.

\bibitem[Sheynin et~al.(2023)Sheynin, Polyak, Singer, Kirstain, Zohar, Ashual, Parikh, and Taigman]{sheynin2023emu}
Shelly Sheynin, Adam Polyak, Uriel Singer, Yuval Kirstain, Amit Zohar, Oron Ashual, Devi Parikh, and Yaniv Taigman.
\newblock Emu edit: Precise image editing via recognition and generation tasks.
\newblock \emph{arXiv preprint arXiv:2311.10089}, 2023.

\bibitem[Sohl-Dickstein et~al.(2015)Sohl-Dickstein, Weiss, Maheswaranathan, and Ganguli]{sohl2015deep}
Jascha Sohl-Dickstein, Eric Weiss, Niru Maheswaranathan, and Surya Ganguli.
\newblock Deep unsupervised learning using nonequilibrium thermodynamics.
\newblock In \emph{International conference on machine learning (ICML)}, pages 2256--2265. PMLR, 2015.

\bibitem[Song et~al.(2020)Song, Meng, and Ermon]{song2020denoising}
Jiaming Song, Chenlin Meng, and Stefano Ermon.
\newblock Denoising diffusion implicit models.
\newblock \emph{arXiv preprint arXiv:2010.02502}, 2020.

\bibitem[Touvron et~al.(2023)Touvron, Lavril, Izacard, Martinet, Lachaux, Lacroix, Rozi{\`e}re, Goyal, Hambro, Azhar, et~al.]{touvron2023llama}
Hugo Touvron, Thibaut Lavril, Gautier Izacard, Xavier Martinet, Marie-Anne Lachaux, Timoth{\'e}e Lacroix, Baptiste Rozi{\`e}re, Naman Goyal, Eric Hambro, Faisal Azhar, et~al.
\newblock Llama: Open and efficient foundation language models.
\newblock \emph{arXiv preprint arXiv:2302.13971}, 2023.

\bibitem[Valevski et~al.(2023)Valevski, Kalman, Molad, Segalis, Matias, and Leviathan]{valevski2023unitune}
Dani Valevski, Matan Kalman, Eyal Molad, Eyal Segalis, Yossi Matias, and Yaniv Leviathan.
\newblock Unitune: Text-driven image editing by fine tuning a diffusion model on a single image.
\newblock \emph{ACM Transactions on Graphics (TOG)}, 42\penalty0 (4):\penalty0 1--10, 2023.

\bibitem[Wang et~al.(2023{\natexlab{a}})Wang, Chen, Ni, Liu, and Liu]{wang2023omni}
Hang Wang, Xuanhong Chen, Bingbing Ni, Yutian Liu, and Jinfan Liu.
\newblock Omni aggregation networks for lightweight image super-resolution.
\newblock In \emph{Proceedings of the IEEE/CVF Conference on Computer Vision and Pattern Recognition}, pages 22378--22387, 2023{\natexlab{a}}.

\bibitem[Wang et~al.(2023{\natexlab{b}})Wang, Zhang, Birsak, and Wonka]{wang2023instructedit}
Qian Wang, Biao Zhang, Michael Birsak, and Peter Wonka.
\newblock Instructedit: Improving automatic masks for diffusion-based image editing with user instructions.
\newblock \emph{arXiv preprint arXiv:2305.18047}, 2023{\natexlab{b}}.

\bibitem[Wang et~al.(2023{\natexlab{c}})Wang, Saharia, Montgomery, Pont-Tuset, Noy, Pellegrini, Onoe, Laszlo, Fleet, Soricut, et~al.]{wang2023imagen}
Su~Wang, Chitwan Saharia, Ceslee Montgomery, Jordi Pont-Tuset, Shai Noy, Stefano Pellegrini, Yasumasa Onoe, Sarah Laszlo, David~J Fleet, Radu Soricut, et~al.
\newblock Imagen editor and editbench: Advancing and evaluating text-guided image inpainting.
\newblock In \emph{Proceedings of the IEEE/CVF conference on computer vision and pattern recognition}, pages 18359--18369, 2023{\natexlab{c}}.

\bibitem[Wang et~al.(2004)Wang, Bovik, Sheikh, and Simoncelli]{wang2004image}
Zhou Wang, Alan~C Bovik, Hamid~R Sheikh, and Eero~P Simoncelli.
\newblock Image quality assessment: from error visibility to structural similarity.
\newblock \emph{IEEE transactions on image processing}, 13\penalty0 (4):\penalty0 600--612, 2004.

\bibitem[Wei et~al.(2018)Wei, Wang, Yang, and Liu]{wei2018deep}
Chen Wei, Wenjing Wang, Wenhan Yang, and Jiaying Liu.
\newblock Deep retinex decomposition for low-light enhancement.
\newblock \emph{arXiv preprint arXiv:1808.04560}, 2018.

\bibitem[Welling and Teh(2011)]{welling2011bayesian}
Max Welling and Yee~W Teh.
\newblock Bayesian learning via stochastic gradient langevin dynamics.
\newblock In \emph{Proceedings of the 28th international conference on machine learning (ICML)}, pages 681--688, 2011.

\bibitem[Wu et~al.(2023{\natexlab{a}})Wu, Yin, Qi, Wang, Tang, and Duan]{wu2023visual}
Chenfei Wu, Shengming Yin, Weizhen Qi, Xiaodong Wang, Zecheng Tang, and Nan Duan.
\newblock Visual chatgpt: Talking, drawing and editing with visual foundation models.
\newblock \emph{arXiv preprint arXiv:2303.04671}, 2023{\natexlab{a}}.

\bibitem[Wu et~al.(2023{\natexlab{b}})Wu, Zhang, Zhang, Chen, Liao, Wang, Li, Sun, Yan, Zhai, et~al.]{wu2023q}
Haoning Wu, Zicheng Zhang, Erli Zhang, Chaofeng Chen, Liang Liao, Annan Wang, Chunyi Li, Wenxiu Sun, Qiong Yan, Guangtao Zhai, et~al.
\newblock Q-bench: A benchmark for general-purpose foundation models on low-level vision.
\newblock \emph{arXiv preprint arXiv:2309.14181}, 2023{\natexlab{b}}.

\bibitem[Wu et~al.(2023{\natexlab{c}})Wu, Duan, Guo, Chai, and Li]{wu2023ridcp}
Rui-Qi Wu, Zheng-Peng Duan, Chun-Le Guo, Zhi Chai, and Chongyi Li.
\newblock Ridcp: Revitalizing real image dehazing via high-quality codebook priors.
\newblock In \emph{Proceedings of the IEEE/CVF Conference on Computer Vision and Pattern Recognition}, pages 22282--22291, 2023{\natexlab{c}}.

\bibitem[Xu et~al.(2024)Xu, Ma, Huang, Lee, and Chai]{xu2024cyclenet}
Sihan Xu, Ziqiao Ma, Yidong Huang, Honglak Lee, and Joyce Chai.
\newblock Cyclenet: Rethinking cycle consistency in text-guided diffusion for image manipulation.
\newblock \emph{Advances in Neural Information Processing Systems}, 36, 2024.

\bibitem[Ye et~al.(2023)Ye, Xu, Xu, Ye, Yan, Zhou, Wang, Hu, Shi, Shi, et~al.]{ye2023mplug}
Qinghao Ye, Haiyang Xu, Guohai Xu, Jiabo Ye, Ming Yan, Yiyang Zhou, Junyang Wang, Anwen Hu, Pengcheng Shi, Yaya Shi, et~al.
\newblock mplug-owl: Modularization empowers large language models with multimodality.
\newblock \emph{arXiv preprint arXiv:2304.14178}, 2023.

\bibitem[Yoon et~al.(2019)Yoon, Jarrett, and Van~der Schaar]{yoon2019time}
Jinsung Yoon, Daniel Jarrett, and Mihaela Van~der Schaar.
\newblock Time-series generative adversarial networks.
\newblock \emph{Advances in neural information processing systems}, 32, 2019.

\bibitem[Yu et~al.(2023)Yu, Yang, Li, Wang, Lin, Liu, Wang, and Wang]{yu2023mm}
Weihao Yu, Zhengyuan Yang, Linjie Li, Jianfeng Wang, Kevin Lin, Zicheng Liu, Xinchao Wang, and Lijuan Wang.
\newblock Mm-vet: Evaluating large multimodal models for integrated capabilities.
\newblock \emph{arXiv preprint arXiv:2308.02490}, 2023.

\bibitem[Zhang et~al.(2019)Zhang, Goodfellow, Metaxas, and Odena]{zhang2019self}
Han Zhang, Ian Goodfellow, Dimitris Metaxas, and Augustus Odena.
\newblock Self-attention generative adversarial networks.
\newblock In \emph{International conference on machine learning}, pages 7354--7363. PMLR, 2019.

\bibitem[Zhang et~al.(2024{\natexlab{a}})Zhang, Mo, Chen, Sun, and Su]{zhang2024magicbrush}
Kai Zhang, Lingbo Mo, Wenhu Chen, Huan Sun, and Yu~Su.
\newblock Magicbrush: A manually annotated dataset for instruction-guided image editing.
\newblock \emph{Advances in Neural Information Processing Systems}, 36, 2024{\natexlab{a}}.

\bibitem[Zhang et~al.(2018)Zhang, Isola, Efros, Shechtman, and Wang]{zhang2018unreasonable}
Richard Zhang, Phillip Isola, Alexei~A Efros, Eli Shechtman, and Oliver Wang.
\newblock The unreasonable effectiveness of deep features as a perceptual metric.
\newblock In \emph{Proceedings of the IEEE conference on computer vision and pattern recognition}, pages 586--595, 2018.

\bibitem[Zhang et~al.(2023{\natexlab{a}})Zhang, Yang, Feng, Qin, Chen, Yu, Chen, Wang, Savarese, Ermon, et~al.]{zhang2023hive}
Shu Zhang, Xinyi Yang, Yihao Feng, Can Qin, Chia-Chih Chen, Ning Yu, Zeyuan Chen, Huan Wang, Silvio Savarese, Stefano Ermon, et~al.
\newblock Hive: Harnessing human feedback for instructional visual editing.
\newblock \emph{arXiv preprint arXiv:2303.09618}, 2023{\natexlab{a}}.

\bibitem[Zhang et~al.(2023{\natexlab{b}})Zhang, Han, Ghosh, Metaxas, and Ren]{zhang2023sine}
Zhixing Zhang, Ligong Han, Arnab Ghosh, Dimitris~N Metaxas, and Jian Ren.
\newblock Sine: Single image editing with text-to-image diffusion models.
\newblock In \emph{Proceedings of the IEEE/CVF Conference on Computer Vision and Pattern Recognition}, pages 6027--6037, 2023{\natexlab{b}}.

\bibitem[Zhang et~al.(2024{\natexlab{b}})Zhang, Zheng, Fang, and Plummer]{zhang2024text}
Zhongping Zhang, Jian Zheng, Zhiyuan Fang, and Bryan~A Plummer.
\newblock Text-to-image editing by image information removal.
\newblock In \emph{Proceedings of the IEEE/CVF Winter Conference on Applications of Computer Vision}, pages 5232--5241, 2024{\natexlab{b}}.

\bibitem[Zhu et~al.(2024)Zhu, Tang, Han, Lu, Zhao, Xing, Wang, and Yin]{zhu2024vislinginstruct}
Dongsheng Zhu, Xunzhu Tang, Weidong Han, Jinghui Lu, Yukun Zhao, Guoliang Xing, Junfeng Wang, and Dawei Yin.
\newblock Vislinginstruct: Elevating zero-shot learning in multi-modal language models with autonomous instruction optimization.
\newblock \emph{arXiv preprint arXiv:2402.07398}, 2024.

\end{thebibliography}
}

\end{document}